%% file: sn-article.tex
\documentclass[sn-mathphys-num]{sn-jnl}

\usepackage{graphicx}%
\usepackage{mathrsfs}%
\usepackage[title]{appendix}%
\usepackage{textcomp}%
\usepackage{manyfoot}%
\usepackage{algorithmicx}%
\usepackage{algpseudocode}%
\usepackage{listings}%

\usepackage{amsmath}
\usepackage{amssymb}
\usepackage{mathtools}
\usepackage{amsthm}

\usepackage[utf8]{inputenc} 
\usepackage[T1]{fontenc}    
\usepackage{hyperref}       
\usepackage{url}            
\usepackage{booktabs}       
\usepackage{amsfonts}       
\usepackage{nicefrac}       
\usepackage{microtype}      
\usepackage[dvipsnames]{xcolor}
\usepackage{natbib}
\usepackage[ruled]{algorithm2e}
\usepackage{float}
\usepackage{makecell}

\usepackage{bm}
\usepackage{multirow}
\usepackage{colortbl}
\usepackage{enumitem}
\usepackage{framed}
\usepackage{tcolorbox}
\usepackage{setspace}
\usepackage{array}
\usepackage{wrapfig}
\usepackage{tabularx}
\usepackage{subcaption}
\usepackage{utfsym}
\newcolumntype{P}[1]{>{\centering\arraybackslash}p{#1}}

\newcommand{\format}[2]{{{}{}}$#1$ ({\color{red}$ \small #2$})}

\definecolor{blip_noun}{rgb}{0.95, 0.56, 0.11}
\definecolor{blip_verb}{rgb}{0.96, 0.78, 0.01}
\definecolor{blip_prep}{rgb}{0.98, 0.87, 0.46}
\definecolor{blip_adj}{rgb}{0.98, 0.81, 0.62}

\definecolor{vit_noun}{rgb}{0.17, 0.38, 0.27}
\definecolor{vit_verb}{rgb}{0.37, 0.72, 0.54}
\definecolor{vit_prep}{rgb}{0.62, 0.83, 0.73}
\definecolor{vit_adj}{rgb}{0.73, 0.87, 0.8}

\usepackage[capitalize,noabbrev]{cleveref}

\renewcommand{\paragraph}[1]{\noindent\textbf{#1}\quad}

\theoremstyle{thmstyleone}%
\newtheorem{theorem}{Theorem}
\theoremstyle{thmstyletwo}%
\newtheorem{remark}{Remark}%

\theoremstyle{thmstylethree}%
\newtheorem{definition}{Definition}%

\begin{document}

\title[Caption in Image]{\centering Image Captions are Natural Prompts \\ for Training Data Synthesis}


\author[1]{\fnm{Shiye} \sur{Lei}}
\equalcont{These authors contributed equally to this work.}

\author[2]{\fnm{Hao} \sur{Chen}}
\equalcont{These authors contributed equally to this work.}

\author[1]{\fnm{Sen} \sur{Zhang}}

\author*[3]{\fnm{Bo} \sur{Zhao}}\email{bo.zhao@sjtu.edu.cn}

\author*[4,1]{\fnm{Dacheng} \sur{Tao}}\email{dacheng.tao@ntu.edu.sg}

\affil[1]{\orgdiv{School of Computer Science}, \orgname{The University of Sydney}}

\affil[2]{\orgdiv{Electrical and Computer Engineering}, \orgname{Carnegie Mellon University}}

\affil[3]{\orgdiv{School of Artificial Intelligence}, \orgname{Shanghai Jiao Tong University}}

\affil[4]{\orgdiv{College of Computing \& Data Science}, \orgname{Nanyang Technological University}}


\abstract{With the rapid development of Artificial Intelligence Generated Content (AIGC), it has become a common practice to train models on synthetic data due to data-scarcity and privacy leakage problems. Owing to massive and diverse information conveyed in real images, it is challenging for text-to-image generative models to synthesize informative training data with hand-crafted prompts. Considering the impressive ability of large generative models, {\it could such models directly synthesize good training images for prediction tasks with proper prompts?} We offer an affirmative response to this question by proposing a simple yet effective method, validated through ImageNet classification. Specifically, we caption each real image with the advanced captioning model to obtain informative and faithful prompts that extract class-relevant information and clarify the polysemy of class names. The image captions and class names are concatenated to prompt generative models for training image synthesis. We show that this simple caption incorporation significantly boosts the informativeness of synthetic data therefore enhancing downstream model generalization. More importantly, besides improvements in data augmentation and privacy preservation, our experiments demonstrate that synthesized images can exceed real data in terms of out-of-distribution robustness. The code is available at \url{https://github.com/LeavesLei/Caption_in_Prompt}.}




\maketitle

\backmatter

\input{sec/introduction}

\input{sec/relatedwork}

\input{sec/preliminary}

\input{sec/method}

\input{sec/experiments}

\input{sec/conclusion}

\begin{appendices}

\input{sec/supplementary}




\end{appendices}


\clearpage

\bibliography{ref}

\end{document}

%% file: sec/introduction.tex
\section{Introduction}
\label{sec:intro}
The past decade has witnessed the huge success of deep learning across various tasks with the support of massive realistic and high-quality training data \citep{russakovsky2015imagenet, he2016deep, lei2021understanding, lei2023understanding}. However, the cost of data collection and annotation has presented prohibitive for the research community in many real-world applications \citep{bansal2022systematic}. In addition, real data inherently involves private and sensitive information that is difficult to eliminate and may thereafter cause public and privacy concerns \citep{shokri2015privacy}. Training models on synthetic data provides a promising recipe to address these two problems in multiple fields. Recently, text-to-image (T2I) foundation models \citep{zhang2023text} ,{\it e.g.}, DALL-E \cite{ramesh2022hierarchical}, Imagen \cite{saharia2022photorealistic}, and Stable Diffusion \cite{rombach2022high}, which are trained on billions of image-text pairs from web-datasets \citep{schuhmann2022laionb} and can generate proper high-quality and real-looking images given textual descriptions, present a natural kit to generate synthetic data for downstream model training \citep{he2023is, sariyildiz2023fake,zhou2023training}. 

\begin{figure}[t]
\centering
\includegraphics[width=0.95\textwidth]{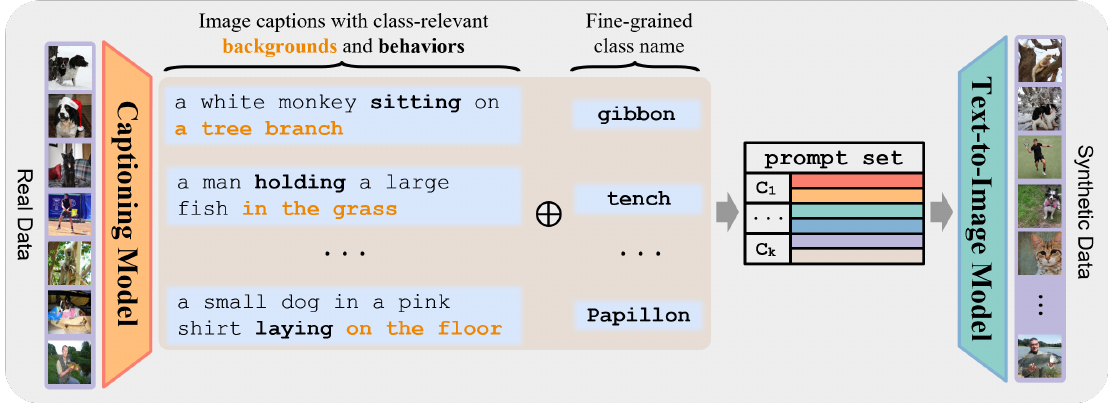}
\caption{The workflow of CiP. The image captions and class names are concatenated  ($\boldsymbol{\oplus}$) to prompt text-to-image model to synthesize informative training samples.}
\label{figure:workflow}
\end{figure}

Several methodologies have been proposed to align the data synthesized by T2I models with real-world data, aiming to enhance the training effect of synthetic datasets, {\it i.e.}, the generalization performance of models exclusively trained on the synthetic data. Fine-tuning T2I models on real data is evidenced to generate more aligned synthetic images \citep{azizi2023synthetic}. Despite promising results, fine-tuning alters model parameters and causes catastrophic forgetting \citep{kirkpatrick2017overcoming,kumari2023multi,smith2024continual}, thereby hurting the knowledge integrity of pre-trained T2I models. To mitigate this issue, inversion-based methods aim to optimize word embeddings to encapsulate the class information \citep{shin2023fill} or conditional vectors to capture the instance information \citep{zhou2023training}. Then informative images can be generated conditioned on corresponding word embeddings or vectors. However, due to direct optimization on real images with the large T2I model, both fine-tune and inversion-based approaches suffer from (1) privacy leakage: real images can be memorized by T2I models during fine-tuning or recovered with optimized conditional vectors; (2) computing inefficiency: for examples, inversion operation in \citep{zhou2023training} requires $84$ seconds to inverse one $256\times 256$ image and thus not scalable to large datasets like ImageNet-1K; and (3) unsatisfied transferability: fine-tuned model or inversed-embeddings/vectors are model-dependent and can not be transferred to other T2I models. Therefore, albeit straightforward, prompt-based methods, which solely modify textual prompts to control the image generation, are more promising to synthesize informative data due to side-stepping aforementioned drawbacks \citep{he2023is, sariyildiz2023fake}; please refer to Table \ref{tab:method comparison} for a holistic comparison.

\begin{table*}[t]
    \centering
    \caption{Comparison between different T2I generation methods.}
    \resizebox{0.85\linewidth}{!}{
    \begin{tabular}{cccccc}
    \toprule
    & Refs 
    & Privacy & Efficiency & Integrity & Transferability \\
    \midrule
    Fine-tune &  \citep{ravuri2019seeing, azizi2023synthetic}  & 
    \usym{2718}  & \usym{2718} & \usym{2718} &  \usym{2718}  \\
    Inversion-based & \citep{zhou2023training,shin2023fill,trabucco2024effective} &  
    \usym{2718} &  \usym{2718} & \usym{2714} & \usym{2718} \\
    Prompt-based & \citep{he2023is,sariyildiz2023fake,bansal2023leaving} & 
    \usym{2714} & \usym{2714} & \usym{2714} & \usym{2714} \\
    \bottomrule
    \end{tabular}
    }
    \label{tab:method comparison}
\end{table*}

Recently, numerous prompt-based methods have been introduced, where synthetic training datasets are constructed by querying T2I models with specific prompts. For example, \citet{sariyildiz2023fake} integrated class definition and multiple backgrounds into textual prompts, and \citet{yuan2022not} augmented the prompt set with CLIP templates. In addition, \citet{he2023is} also employed large language models (LLMs) to generate diverse prompts. However, according to \citep{shin2023fill}, the basic template, {\it i.e.}, \texttt {a photo of \{class name\}}, performs the best among these strategies. This invokes us to rethink {\it whether prompt design advances informative data synthesis for downstream model training}, a fundamental step in identifying the potential of T2I models for training image synthesis.

In this paper, we offer an affirmative response to this question. Concretely, we investigate whether the class-relevant textual information in prompts can be decoded by T2I models and induce class-relevant visual information for learning prediction tasks. A simple and training-free framework, {\it \textbf{C}aption \textbf{i}n \textbf{P}rompt} (CiP), is proposed to construct class-relevant and informative prompts and achieve computationally efficient generation of informative synthetic training samples. Specifically, CiP generates high-quality textual prompts by leveraging the off-the-shelf image captioning models \citep{kumar2022imagecaptioning,li2023blip} to caption real images. The captions and class names are then concatenated to constitute the prompts. The overall pipeline of CiP is presented in Figure \ref{figure:workflow}. More diverse and informative training samples are consequently generated, as shown in Table \ref{tab:real_blip}. Since CiP is training-free, the proposed method is scalable to arbitrary-size datasets.

\begin{table}[t]
\footnotesize
\setlength\tabcolsep{1pt}
    \centering
    \caption{Images captions of real images and corresponding synthetic (Syn) images. Last column shows the syn images generated by basic prompts (BSyn Images).}
    \begin{tabular}{p{0.8cm}|cp{8cm}c|c}
    \toprule
    &  Real &  \multicolumn{1}{c}{ Captions} &  Syn &  BSyn \\
    \specialrule{1pt}{0.2\jot}{0.1pc}
    \multirow{11}{*}{\rotatebox{90}{\thead{English Springer Spaniel \\ (n02102040)}}} & \raisebox{-0.7\totalheight}{\includegraphics[width=0.08\textwidth, height=0.08\textwidth]{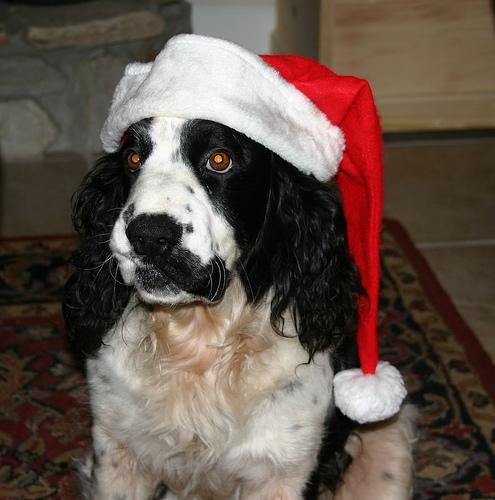}} &  \multirow{3}{8.0cm}{\centering \texttt{ a dog wearing a santa hat}}&  \raisebox{-0.7\totalheight}{\includegraphics[width=0.08\textwidth]{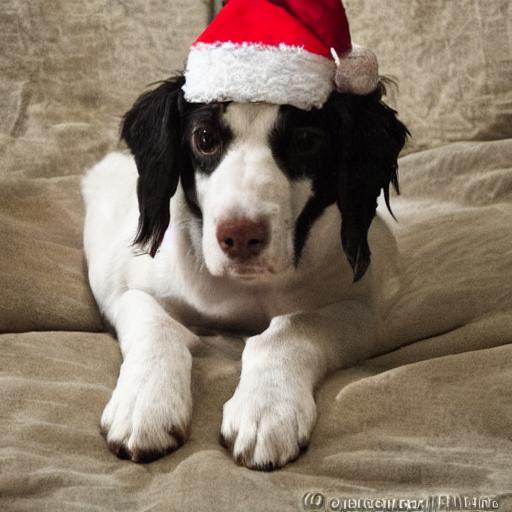}} &  \raisebox{-0.7\totalheight}{\includegraphics[width=0.08\textwidth]{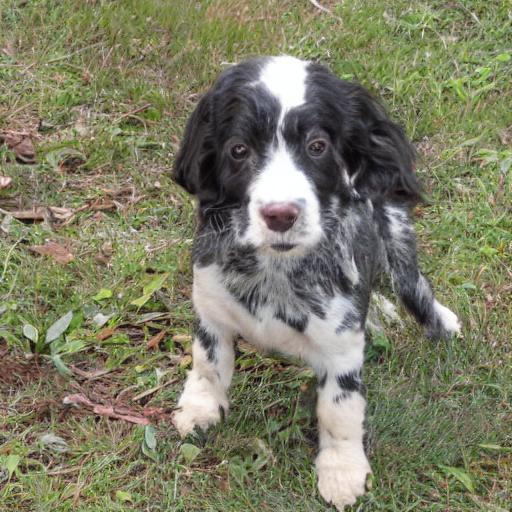}}\\
    & \raisebox{-0.9\totalheight}{\includegraphics[width=0.08\textwidth, height=0.08\textwidth]{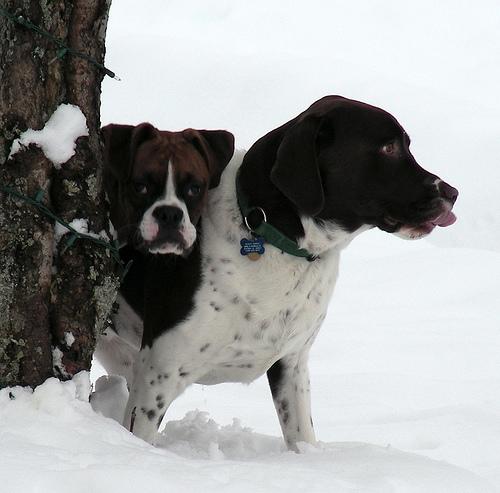}} &  \multirow{4}{8.0cm}{\centering\texttt{ two dogs standing next to a tree in the snow}} &  \raisebox{-0.9\totalheight}{\includegraphics[width=0.08\textwidth]{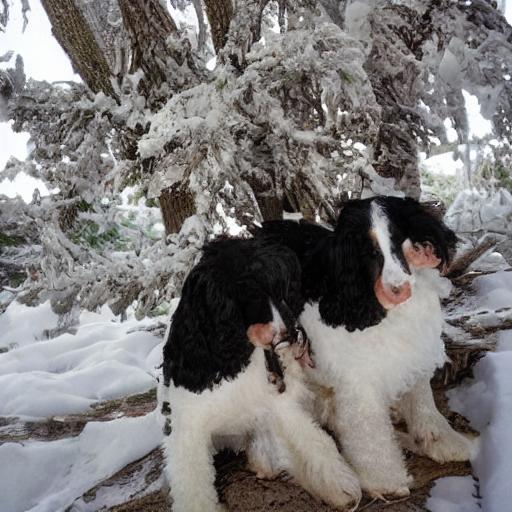}} &  \raisebox{-0.9\totalheight}{\includegraphics[width=0.08\textwidth]{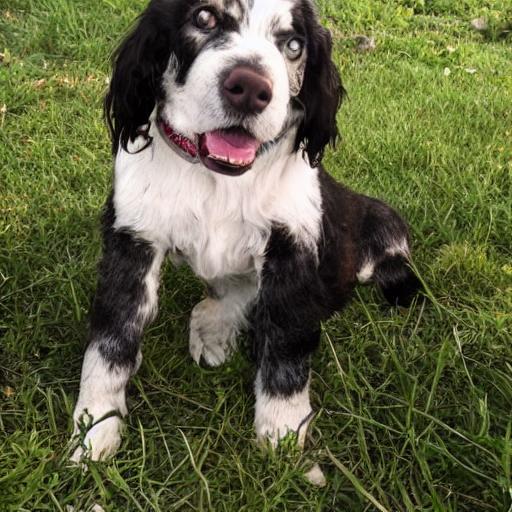}}\\
    & \raisebox{-0.9\totalheight}{\includegraphics[width=0.08\textwidth, height=0.08\textwidth]{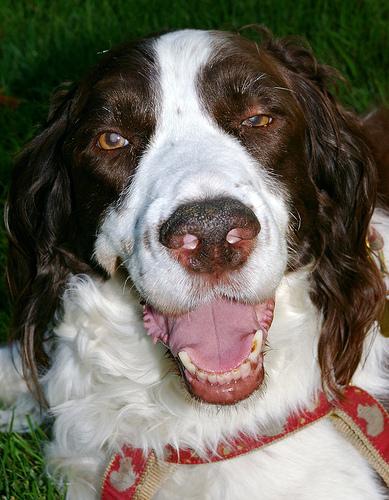}} &  \multirow{4}{8.0cm}{\centering\texttt{ a brown and white dog with a red collar}}
 &  \raisebox{-0.9\totalheight}{\includegraphics[width=0.08\textwidth]{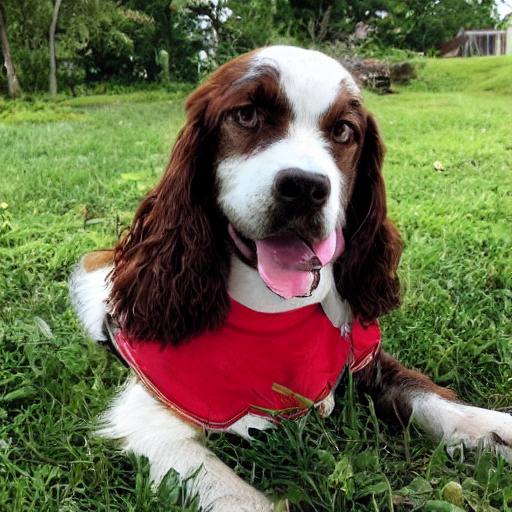}} &  \raisebox{-0.9\totalheight}{\includegraphics[width=0.08\textwidth]{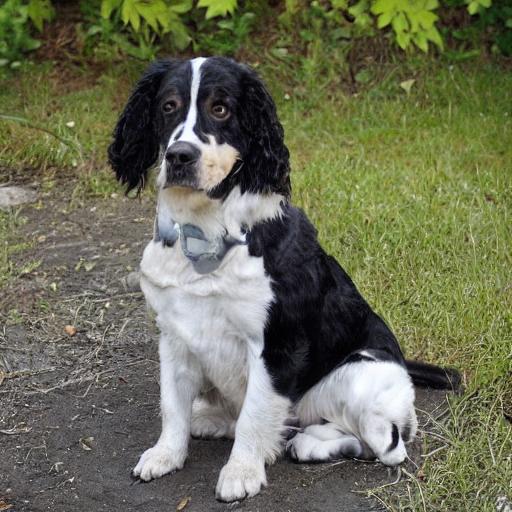}}\\
 \specialrule{0.5pt}{0.2\jot}{0.1pc}

 \multirow{10}{*}{\rotatebox{90}{\thead{Couch \\ (n04344873)}}} & \raisebox{-0.5\totalheight}{\includegraphics[width=0.08\textwidth, height=0.08\textwidth]{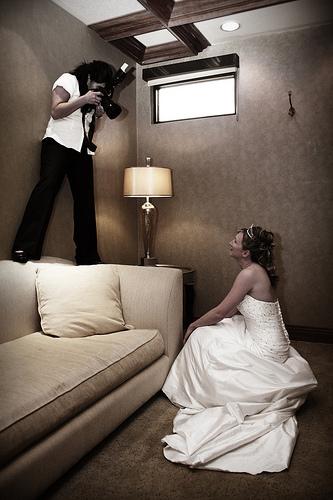}} &  \multirow{2}{8.0cm}{\centering\texttt{ a woman in a wedding dress
}} &  \raisebox{-0.5\totalheight}{\includegraphics[width=0.08\textwidth]{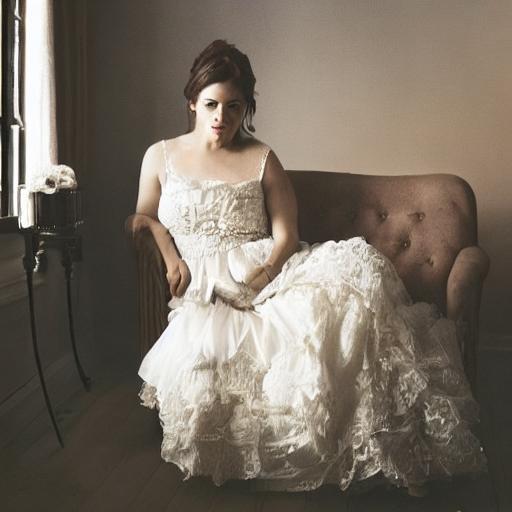}} &  \raisebox{-0.5\totalheight}{\includegraphics[width=0.08\textwidth]{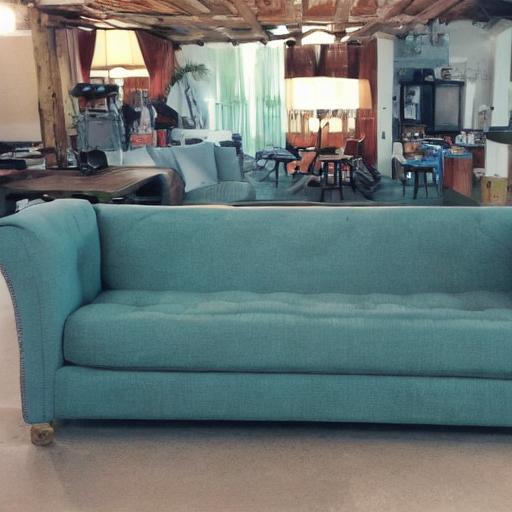}}\\
    & \raisebox{-0.9\totalheight}{\includegraphics[width=0.08\textwidth, height=0.08\textwidth]{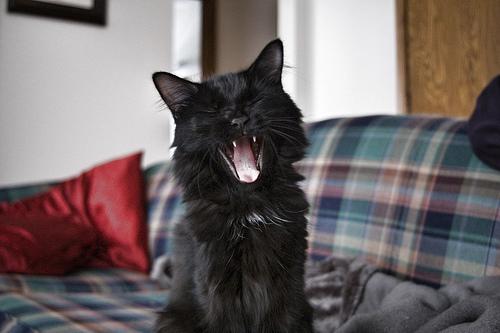}} &  \multirow{4}{8.0cm}{\centering\texttt{ a black cat yawning on a couch
}} &  \raisebox{-0.9\totalheight}{\includegraphics[width=0.08\textwidth]{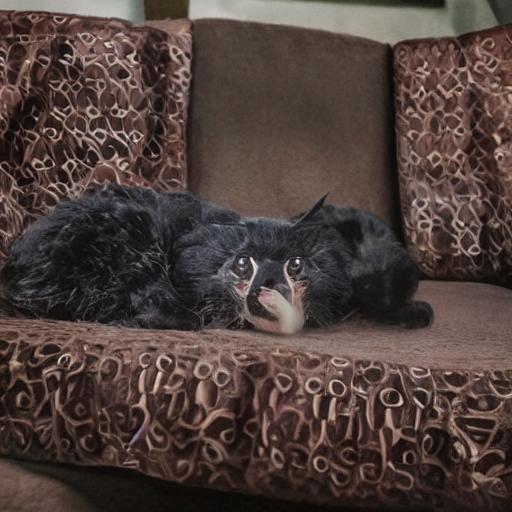}} &  \raisebox{-0.9\totalheight}{\includegraphics[width=0.08\textwidth]{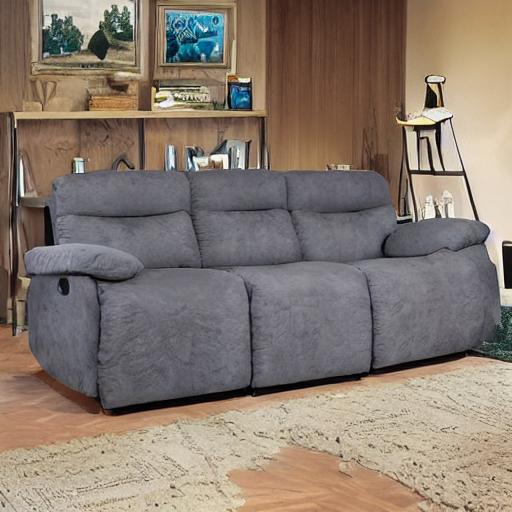}}\\
    & \raisebox{-0.9\totalheight}{\includegraphics[width=0.08\textwidth, height=0.08\textwidth]{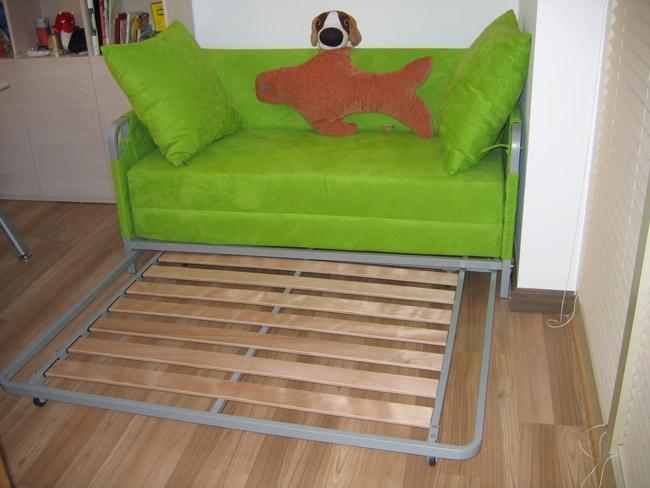}} &  \multirow{4}{8.0cm}{\centering\texttt{ a green couch with a stuffed animal on it}}
 &  \raisebox{-0.9\totalheight}{\includegraphics[width=0.08\textwidth]{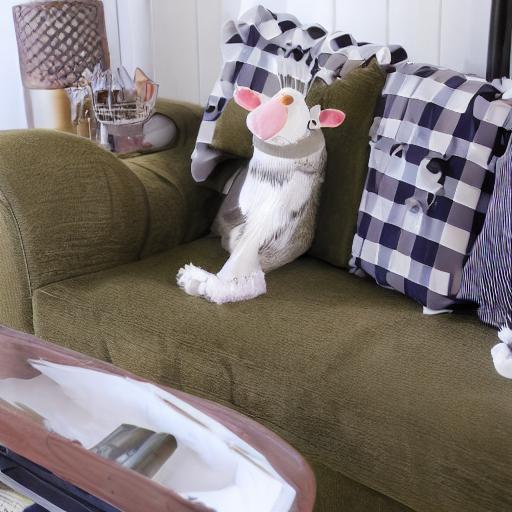}} &  \raisebox{-0.9\totalheight}{\includegraphics[width=0.08\textwidth]{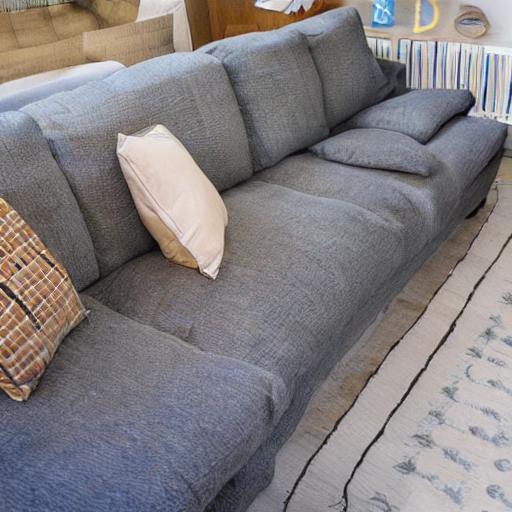}}\\
 \specialrule{0.5pt}{0.2\jot}{0.1pc}

  \multirow{11}{*}{\rotatebox{90}{\thead{Baseball \\ (n02799071)}}} & \raisebox{-0.7\totalheight}{\includegraphics[width=0.08\textwidth, height=0.08\textwidth]{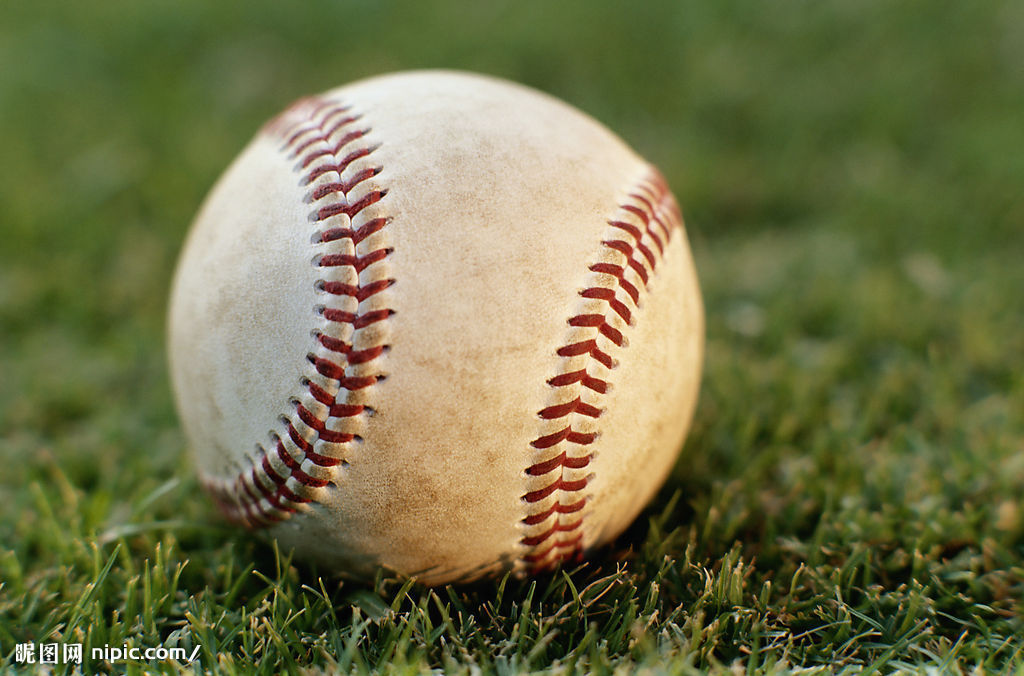}} &  \multirow{3}{8.0cm}{\centering\texttt{ a baseball sitting on the grass in the field}} &  \raisebox{-0.7\totalheight}{\includegraphics[width=0.08\textwidth]{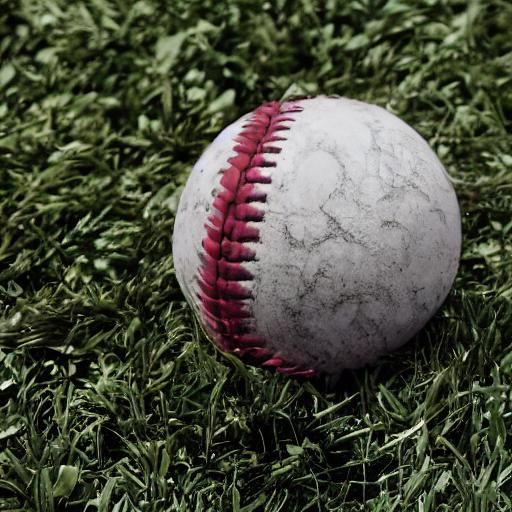}} &  \raisebox{-0.7\totalheight}{\includegraphics[width=0.08\textwidth]{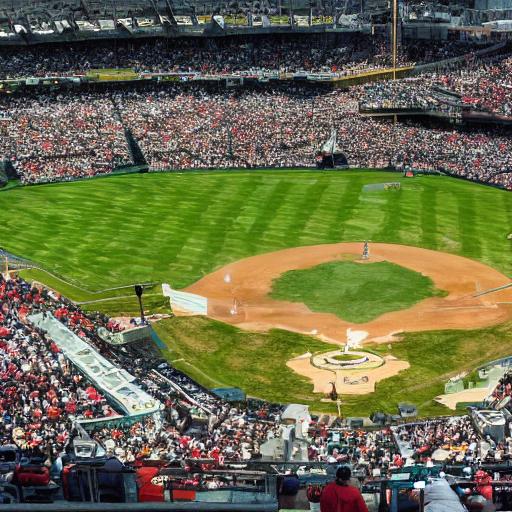}}\\
    & \raisebox{-0.9\totalheight}{\includegraphics[width=0.08\textwidth, height=0.08\textwidth]{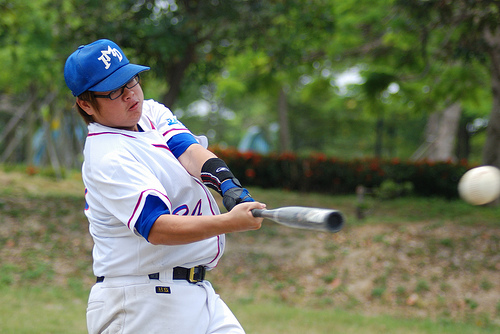}} &  \multirow{4}{8.0cm}{\centering\texttt{ a baseball player swinging a bat at a ball}} &  \raisebox{-0.9\totalheight}{\includegraphics[width=0.08\textwidth]{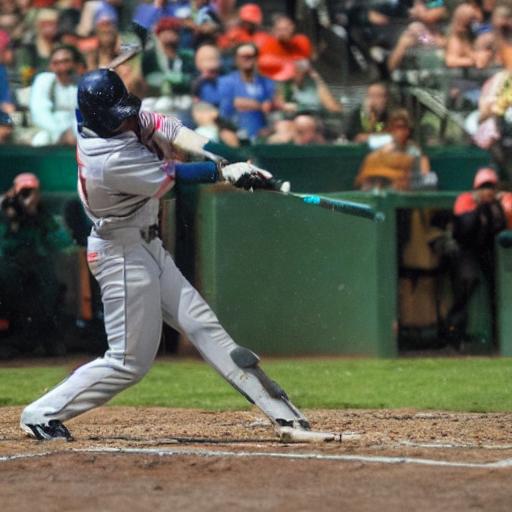}} &  \raisebox{-0.9\totalheight}{\includegraphics[width=0.08\textwidth]{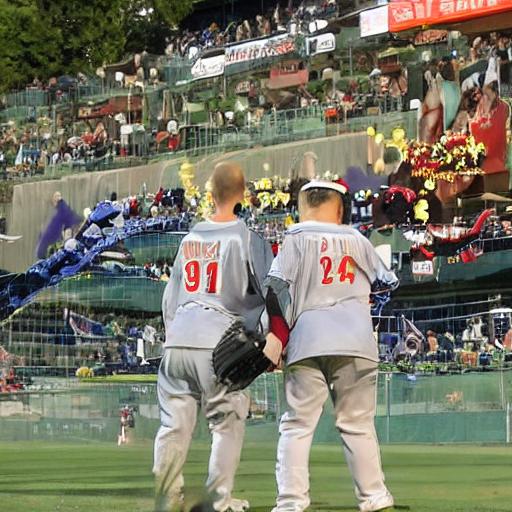}}\\
    & \raisebox{-0.9\totalheight}{\includegraphics[width=0.08\textwidth, height=0.08\textwidth]{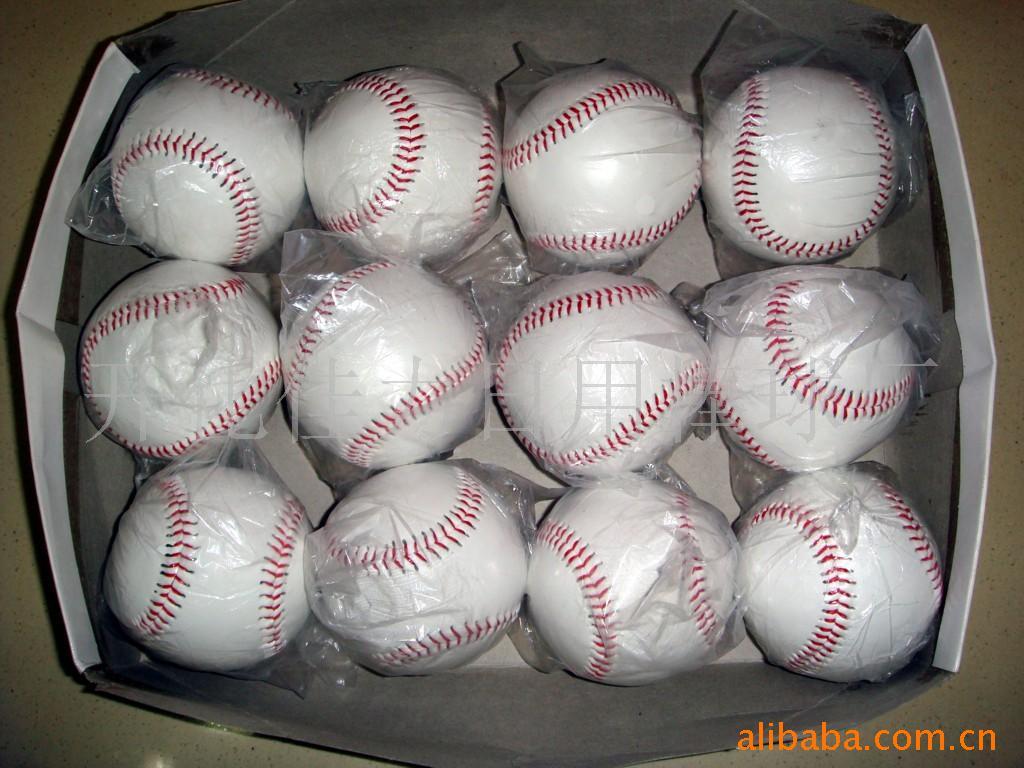}} &  \multirow{4}{8.0cm}{\centering\texttt{ a box of baseballs in plastic wrap}}
 &  \raisebox{-0.9\totalheight}{\includegraphics[width=0.08\textwidth]{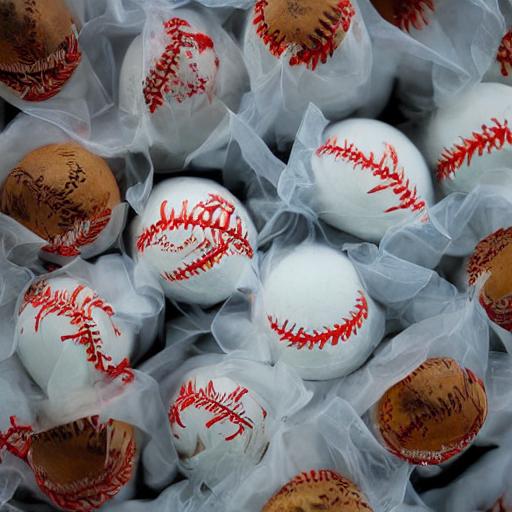}} &  \raisebox{-0.9\totalheight}{\includegraphics[width=0.08\textwidth]{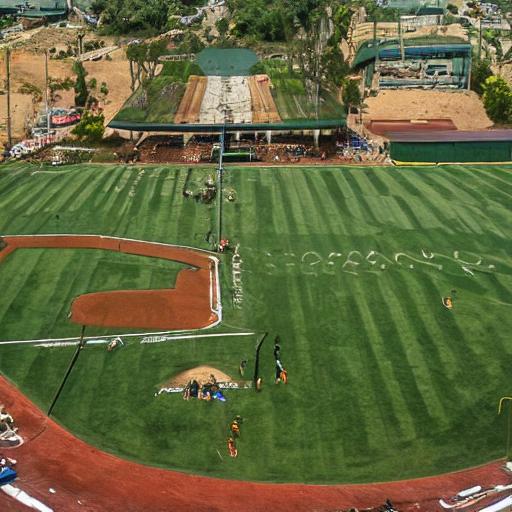}}\\
    \bottomrule
    \end{tabular}
    \label{tab:real_blip}
\end{table}

Extensive experiments on ImageNette \citep{imagewang}, ImageNet-100 \citep{tian2020contrastive}, and ImageNet-1K \citep{russakovsky2015imagenet} demonstrate that our method can improve the classification accuracy of models trained on synthetic data by a substantial margin of around ${10}\%$. This remarkable gain reveals that proper prompt design can easily elevate the training effect of synthetic data. Moreover, models trained on CiP-induced data also achieve higher out-of-distribution (OoD) accuracy {by $5.70\%$ in ImageNet-R \citep{hendrycks2021many} and $2.39\%$ in ImageNet-A \citep{hendrycks2021natural}} than real data, which is firstly observed in ImageNet-1K by solely training models on T2I synthetic data. From the privacy perspective, our method demonstrates complete immunity to membership inference attack (MIA) by achieving the MIA accuracy of $0.5$, {\it i.e.}, random guess. In addition, the synthetic data induced by CiP can further boost generalization when trained with real data in the data augmentation setting. Further empirical analysis suggests that this improvement brought by CiP can be attributed to class-relevant information extraction (behaviors and backgrounds) from real data and clarification of polysemy. Our contributions can be summarized as:
\begin{itemize}[leftmargin=5mm]
    \item We investigate the training effect of synthetic data from the perspective of prompt design, and a training-free method of CiP is proposed to prompt T2I models to synthesize informative data;
    \item Extensive experiments conducted on multiple datasets including ImageNet-1K, verify that our method can synthesize high-quality images that achieve superior performance on boosting model generalization, OoD robustness, privacy protection, and data augmentation;
    \item Multiple ablation studies with followed post-hoc analysis are provided to illustrate the effectiveness of our methods, which  provides guidelines for prompting T2I models to synthesize informative training data.
\end{itemize}

%% file: sec/relatedwork.tex
\section{Related Work}

\paragraph{Text-to-Image Models} 
In the past decade, generative adversarial networks (GANs) \citep{goodfellow2014generative,arjovsky2017wasserstein,brock2018large,karras2018progressive} are the most successful and popular generative models and can synthesize high-resolution images \citep{ledig2017photo,xie2018tempogan} with specific conditions including texts \citep{reed2016generative,zhang2017stackgan} and images \citep{isola2017image,zhu2017unpaired}. However, there also exists some tricky problems in GAN, like lack of generation diversity \citep{razavi2019generating,nichol2021improved,nash2021generating} and non-stationary training \citep{brock2018large,miyato2018spectral,brock2016neural}. Recently, diffusion probabilistic models \citep{ho2020denoising}, which denoise images step-by-step, such as GLIDE \cite{nichol2021glide}, Imagen \cite{saharia2022photorealistic}, and Stable Diffusion \cite{rombach2022high}, have achieved the new state of the art in photo-realistic image synthesis \citep{dhariwal2021diffusion}. Different from GANs mostly trained on data with respect to ({\it w.r.t.}) specific classes or domains, diffusion models are normally pre-trained on billion-scale image-text pair datasets like LAION \citep{schuhmann2022laionb}, thereby entailing general knowledge about the open world. This allows researchers to extract specific knowledge or information into the generated image by enquiring diffusion models with proper textual prompts.

\paragraph{Synthetic Training Dataset Generation} 
Many works have explored the usage of diffusion models generated images as the training set \citep{he2023is,sariyildiz2023fake,zhou2023training,tian2023stablerep,tian2024learning} or augmenting original real datasets \citep{yuan2022not,bansal2023leaving,azizi2023synthetic,trabucco2024effective,dunlap2023diversify}. 
Extensive experiments in \cite{he2023is} show that synthetic images can remarkably boost zero-shot and few-shot learning, while multiple times of synthetic images are needed to achieve comparable pre-training results to real images. To reduce the semantic ambiguity and enrich diversity, \citet{sariyildiz2023fake} prompt diffusion models with the WordNet information and specific backgrounds. \citet{azizi2023synthetic} fine-tune Imagen model on ImageNet-1K to synthesize images for data augmentation. The above methods all prompt the diffusion models with class-level information, {\it e.g.}, names and WordNet information, which is inferior {\it w.r.t.} diversity. In contrast, we use instance-level prompts, {\it i.e.}, image captions, to prompt T2I models and generate diverse images to achieve better training effects. 

Recently, a few attention has been directed towards fully synthetic CLIP training \citep{tian2024learning,hammoud2024synthclip}. In these approaches, synthetic captions are initially generated by LLMs to prompt T2I models in synthesizing images. The resulting caption-image pairs are subsequently used to train CLIP through contrastive learning. In contrast, our work emphasizes supervised learning within the context of classification tasks, allowing for a more focused exploration of the training effect of synthetic images without delving extensively into the caption generation process.

The concurrent related work \citep{dunlap2023diversify} presents the most similar work to ours, which also involves captioning models to assist image generation. Specifically, the authors summarize {domains}, {\it e.g.}, location and weather, from real image captions using LLMs and generate a few general domain descriptions as the suffix to edit real images through T2I models for data augmentation. 
However, it is difficult to summarize general domains from a coarse-grained dataset like ImageNet-1K. Different from \citep{dunlap2023diversify}, we leverage captions of real images to synthesize more diverse images instead of augmenting real data via image editing. 

%% file: sec/preliminary.tex
\section{Preliminaries}
\label{sec:preliminary}

\paragraph{Notations}
We denote the dataset by $\mathcal{T}=\{(\boldsymbol{x}_i, y_i)\}_{i=1}^m$, where $\boldsymbol{x}_i \in \mathcal{X} \subset \mathbb{R}^d$, $d$ is the input dimension, 
$y_i\in \mathcal{K} = \{1,\dots, k\}$, and $k=|\mathcal{K}|$ is the number of classes. 
We assume that $(\boldsymbol{x}_i, y_i)$ are independent and identically distributed (i.i.d.) random variables drawn from the data generating distribution $\mathcal{D}=\mathcal{X} \times \mathcal{K}$, where $\mathcal{X} = \mathbf{P}(\boldsymbol{x})$ is called data distribution\footnote{The data distribution $\mathcal{X}=\mathbf{P}(\boldsymbol{x})$ is also referred to as a covariate distribution in the transfer learning literature.}. 
As the T2I model generates images conditioning on textual prompts, we denote the T2I model as $f(\mathbf{z}|\mathbf{t}): \mathbb{R}^n \rightarrow \mathbb{R}^d$ that draws the random latent vector $\mathbf{z}\sim \mathcal{N}(\mathbf{0}, \mathbf{I})$ as the input and generates the synthetic image $\boldsymbol{s}\in \mathbb{R}^d$ conditioning on the given prompt $\mathbf{t}$. Due to the randomness of $\mathbf{z}$, the distribution of generated $\boldsymbol{s}$ is denoted as $\mathbf{P}(\boldsymbol{s}| \mathbf{t}, f)$, which is simplified as $\mathbf{P}(\boldsymbol{s}|\mathbf{t})$ by omitting the generative model $f$. To induce class-conditional images from $f$, the prompt $\mathbf{t}$ is required to possess the class information ({\it e.g.}, class names), thereby $y = \mathcal{I}(\mathbf{t})$ by a predefined mapping $\mathcal{I}$. Therefore, given a prompt $\mathbf{t}_0$, we can generate the synthetic data point $(\boldsymbol{s_0}, y_0)$ through $\boldsymbol{s}_0 = f(\mathbf{z} | \mathbf{t}_0)$ and $y_0 = \mathcal{I}(\mathbf{t}_0)$, and the generated synthetic dataset is denoted as $\mathcal{S} = \{(\boldsymbol{s}_i, y_i)\}_{i=1}^m$.

As for training the downstream classification model $\mathcal{M}_{\boldsymbol{\theta}}: \mathbb{R}^d \rightarrow \mathbb{R}^k$, $\boldsymbol{\theta} = \texttt{alg}(\mathcal{S})$ denotes the learned parameters returned by leveraging training algorithms ({\it e.g.}, SGD \citep{robbins1951stochastic}) on the dataset $\mathcal{S}$, and thus the well-trained model is denoted as $\mathcal{M}_{\texttt{alg}(\mathcal{S})}$. The expected risk {\it w.r.t.} $\mathcal{D}$ is then formulated as 
\begin{equation}
    \mathcal{R}_{\mathcal{D}}\left(\texttt{alg}\left(\mathcal{S}\right)\right) = \mathbb{E}_{(\boldsymbol{x},y) \sim \mathcal{D}}\left[\mathbb{I} \left( \mathcal{M}_{\texttt{alg}(\mathcal{S})}\left( \boldsymbol{x} \right) \neq y  \right) \right].
\end{equation}
The expected classification accuracy can be conjugated as $\texttt{ACC}_{\mathcal{D}}\left(\texttt{alg}\left(\mathcal{S}\right)\right) = 1 - \mathcal{R}_{\mathcal{D}} \left( \texttt{alg}\left(\mathcal{S}\right)\right)$. As $\mathcal{D}$ is unknown, $\mathcal{R}_{\mathcal{D}}$ or $\texttt{ACC}_{\mathcal{D}}$ is usually estimated on a held-out test set. 

\subsection{Problem Setup}
For a prompt set $\mathcal{C}=\{\mathbf{t}_i\}_{i=1}^{|\mathcal{C}|}$ consisting of textual prompts, the generative data distribution of $\mathcal{C}$ is defined as follows.
\begin{definition}[Prompt-induced data distribution]
Given a T2I model $f(\mathbf{z}|\mathbf{t})$, synthetic data $\boldsymbol{s}$ are generated by uniformly drawing $\mathbf{t}$ from the prompt set $\mathcal{C}=\{\mathbf{t}_i\}_{i=1}^{|\mathcal{C}|}$. The prompt-induced data distribution of $\boldsymbol{s}$ {\it w.r.t.} $\mathcal{C}$ is defined as $ \mathcal{X}_\mathcal{C} = \sum_{i=1}^{|\mathcal{C}|} \frac{1}{|\mathcal{C}|}\mathbf{P}(\boldsymbol{s}|\mathbf{t}_i)$.
\end{definition}

As the synthetic dataset $\mathcal{S}$ are drawn from $\mathcal{D}_\mathcal{C} = \mathcal{X}_\mathcal{C} \times \mathcal{K}$, we aim to generate the prompt set $\mathcal{C}=\{\mathbf{t}_i\}_{i=1}^{|\mathcal{C}|}$ such that $\mathcal{S}$ has a good {\it training effect}, {\it i.e.}, downstream models trained on $\mathcal{S}$ has a satisfied performance {\it w.r.t.} $\mathcal{D}$. In this paper, we use {\it classification accuracy}
$\texttt{ACC}_{\mathcal{D}}\left(\texttt{alg}\left(\mathcal{S}\right)\right)$ 
to evaluate the training effect of $\mathcal{S}$. This problem of generating $\mathcal{C}$ to induce highly-informative $\mathcal{S}$ can thus be formulated as follows:
\begin{align}
    \mathcal{C}^\ast = \arg\min_{\mathcal{C}} \mathbb{E}_{\mathcal{S}\sim \mathcal{X}_\mathcal{C} \times \mathcal{K}}\left[ \mathcal{R}_{\mathcal{D}}\left(\texttt{alg}\left(\mathcal{S}\right)\right)\right] \quad \text{s.t.} \quad |\mathcal{C}| = |\mathcal{S}| = |\mathcal{T}|.
\end{align}
Here we constrain the size of $|\mathcal{C}|$ and $|\mathcal{S}|$ to focus on optimizing the content of textual prompt $\mathbf{t}$.

\subsection{Theoretical Analysis}
\label{sec:theoretical analysis}
With the original distribution $\mathcal{D}$ and the prompt-induced distribution $\mathcal{D}_{\mathcal{C}}$, if the labeling functions $\mathbf{P}(y|\boldsymbol{x})$ are the same for $\mathcal{D}$ and $\mathcal{D}_{\mathcal{C}}$, {\it}  we can derive the following bound. 
\begin{theorem}
\label{thm:1}
    Given distributions $\mathcal{D}=\mathcal{X}\times \mathcal{K}$ and $\mathcal{D}_{\mathcal{C}} =\mathcal{X}_{\mathcal{C}}\times \mathcal{K}$ and $\mathbf{P}_{\mathcal{D}}(y|\boldsymbol{x}) = \mathbf{P}_{\mathcal{D}_{\mathcal{C}}}(y|\boldsymbol{x})$, the training dataset $\mathcal{S}$ is i.i.d. drawn from $\mathcal{D}_{\mathcal{C}}$, then we have
    \begin{equation}
    \mathcal{R}_{\mathcal{D}}(\texttt{alg}(\mathcal{S})) \leq \mathcal{R}_{\mathcal{D}_{\mathcal{C}}}(\texttt{alg}(\mathcal{S})) + d(\mathcal{X}, \mathcal{X}_{\mathcal{C}}),
\end{equation}
where $d(\mathcal{X},  \mathcal{X}_{\mathcal{C}}) = \mathbb{E}_{\boldsymbol{x}}\left[|\mathbf{P}_{\mathcal{X}}\left(\boldsymbol{x}\right) - \mathbf{P}_{\mathcal{X}_{\mathcal{C}}}\left(\boldsymbol{x}\right) |\right]$.
\end{theorem}
We leave the proof in Appendix \ref{app:proof of thm 1}, and a remark is presented with the above theorem.
\begin{remark}
    The expected risk $\mathcal{R}_{\mathcal{D}}(\texttt{alg}(\mathcal{S}))$ is bounded by (1) the classification complexity of $\mathcal{D}_{\mathcal{C}}$, which is entailed in $\mathcal{R}_{\mathcal{D}_\mathcal{C}}(\texttt{alg}(\mathcal{S}))$; and (2) the distance between $\mathcal{X}$ and $\mathcal{X}_\mathcal{C}$.
\end{remark}
To improve the training effect of the synthetic dataset $\mathcal{S}$ thereby reducing the risk $\mathcal{R}_{\mathcal{D}}(\texttt{alg}(\mathcal{S}))$, the above theorem and remark suggest that the prompt set should: (1) provide distinguishable features ({\it e.g.}, class-relevant objects, actions, and backgrounds) across different classes for easy classification, which indicates class-relevant prompts with less class-agnostic information such as uniform backgrounds for all classes; and (2) capture the information of the original distribution $\mathcal{D}$ as much as possible.

%% file: sec/method.tex
\section{Synthesis with Grounded Prompts}
\label{sec:methods}
Different from many works that improve image aesthetics via prompt engineering \citep{hao2023optimizing}, in this paper we investigate the influence of prompts on the {\it training effect of synthetic data} generated from large T2I models. Most existing works exclusively employ class names {\it w.r.t.} the real dataset $\mathcal{T}$ as prompts to generate corresponding synthetic dataset $\mathcal{S}$ \citep{sariyildiz2023fake,bansal2023leaving}. Researchers also enrich prompts with LLMs or ad-hoc phrases (\textit{e.g.} multiple templates and backgrounds, class explanation) \citep{he2023is,sariyildiz2023fake,shin2023fill}.
However, both strategies hardly perform better than naive class name prompts in inducing synthetic data to train downstream models, which raises the question: {\it could we design prompts to generate more informative synthetic data from large pre-trained T2I models?} Addressing this inquiry is the fundamental step in elucidating the capacity of T2I models for training image synthesis.

\begin{algorithm}[t]
\SetKwInOut{KwIn}{Input}
\SetKwInOut{KwOut}{Output}
\SetKw{KwBy}{by}
\caption{Caption in Prompt}
\label{alg:real data}
\KwIn{Image captioning model \texttt{CM},  text-to-image model \texttt{T2I}, target dataset $\mathcal{T}$}
\KwOut{Synthetic dataset $\mathcal{S}$}
$\mathcal{C} \gets \{\}$, $\mathcal{S} \gets \{\}$\;
$\triangleright$ Image Captioning.\\
\For{$(\boldsymbol{x}_i, y_i)$ \textbf{\text{in}} $\mathcal{T}$}{
$\mathbf{t}_i \gets \texttt{CM}(\boldsymbol{x}_i)$ \\
$\mathbf{t}_i \gets \texttt{concate}(y_i, t_i)$ \\
$\mathcal{C}.\texttt{append}(\mathbf{t}_i)$
}
$\triangleright$ Image Synthesizing.\\
\For{$\mathbf{t}_i$ \textbf{\text{in}} $\mathcal{C}$}{
$\boldsymbol{x}_i \gets \texttt{T2I}(\mathbf{t}_i)$ \\
$\mathcal{S}.\texttt{append}\left(\left(\boldsymbol{s}_i, y_i\right)\right)$
}
\end{algorithm}

Since class-relevant information, including unique foregrounds, backgrounds, behaviors, {\it etc.}, is necessary to learning proper representation for different classes, the critical step in answering this question is to explore whether the class-relevant textual information can be decoded by T2I models and induce corresponding visual information. 
We propose {\it \textbf{C}aption \textbf{i}n \textbf{P}rompt} (CiP) to construct data-corresponded prompts that contain class-relevant textual information in terms of real data. Concretely, CiP first extracts meaningful captions from the real dataset $\mathcal{T}$. 
Then a prompt set is correspondingly constructed based on image captions and fed into T2I models for informative synthetic image generation. Through image captions, the prior knowledge of the real dataset with foregrounds and backgrounds can be naturally injected into the synthetic dataset. The pseudo code of CiP is presented in Algorithm \ref{alg:real data}, and the method can be partitioned into three parts as follows.
\begin{itemize}[leftmargin=5mm]
    \item \textbf{Caption generation.} Given the original image dataset $\mathcal{T}=\{(\boldsymbol{x}_i, y_i)\}_{i=1}^m$, we 
    use the image captioning model \texttt{CM} to generate captions $\mathbf{t}_i$ for each image $\boldsymbol{x}_i$ in $\mathcal{T}$, {\it i.e.}, $\mathbf{t}_i = \texttt{CM}(\boldsymbol{x}_i)$, and obtain the caption set $\mathcal{C}=\{\mathbf{t}_i\}_{i=1}^m$; please refer to Table \ref{tab:real_blip} 
    for caption examples.

    \item \textbf{Prompt set construction.} While generated image captions possess vivid details of each original image, synthesizing images with such captions may not be ideal training samples. Firstly, the text-to-image model is not able to distinguish the foreground and background objects in a caption. Thus, the discriminative class information may be suppressed by the other background concepts. Secondly, image captioning may lose fine-grained class information, and lead to coarse-grained image synthesis. For example, images of ``tench'' and ``goldfish'' could be both captioned as ``fish'' and thus general ``fish'' are synthesized. To alleviate these problems, we prefix the class name to the caption as ``\texttt{A photo of \{class\}, \{image caption\}}'' for prompt set construction, so that the discriminative fine-grained class information and diverse background components are integrated into the prompts, bringing informative synthetic images.

    \item \textbf{Synthetic dataset generation.} After prefixing class names to captions, we can get the diverse prompt set belonging to fine-grained class. Then, the constructed prompt set is fed into T2I models for generating the synthetic dataset $\mathcal{S}$. For each prompt, one image is accordingly generated, and the final synthetic dataset $\mathcal{S}$ has the same sample size as $\mathcal{T}$. The synthetic set can be easily enlarged by combining each prompt with multiple random noises $\mathbf{z}$ when prompting T2I models. 
\end{itemize}

%% file: sec/experiments.tex
\section{Evaluation}
\label{sec:evaluation}

\begin{table*}[t]
\scriptsize
    \centering
    \renewcommand{\arraystretch}{1}
    \caption{Top-1 classification accuracy ($\%$) of ResNet-50 trained on synthetic ImageNette (INette) and ImageNet-100 (INet-100). Best results for each guidance scale value and for each method are marked with \textbf{bold scores} and \fcolorbox{white}{violet!15}{purple background}, respectively. {\color{orange} Orange-colored} scores denote performances of zero-shot CiP higher than basic prompts.}
    \resizebox{\linewidth}{!}{
    \begin{tabular}{c ccccc ccccc c}
    \toprule
    & \multirow{2}{*}{Methods} & \multicolumn{9}{c}{Guidance Scale} & \multirow{2}{*}{Real data} \\
    \cmidrule{3-11}
    & & $1$ & $1.5$ & $2$ & $2.5$ & $3$ & $4$ & $5$ & $6$ & $7.5$  & \\
    \specialrule{1pt}{0.2\jot}{0.1pc}
    \cellcolor{blue!15} & Basic prompts & $65.2$ & \cellcolor{violet!15}{$68.4$} & $64.8$ & $66.6$ & $62.2$ & $57.2$ & $55.2$ & $50.8$ & $45.8$ & \multirow{4}{*}{91.4}\\
    \cellcolor{blue!15}  & CiP (zero-shot) & \color{orange}$65.8$ & \cellcolor{violet!15}{$67.8$} & \color{orange}{$65.6$} & $66.4$ & $62.0$ & \color{orange}$61.6$ & \color{orange}$56.0$ & \color{orange}$55.8$ & \color{orange}$49.2$ \\
    \cellcolor{blue!15} & CiP (ViT-GPT2) & $71.0$ & \cellcolor{violet!15}{$77.0$} & $72.0$ & $73.2$ & $71.8$ & $69.4$ & $66.4$ & $59.4$ & $57.2$ & \\
    \multirow{-4}{*}{\cellcolor{blue!15}\rotatebox{90}{INette}} & CiP (BLIP-2) & \bm{$77.4$} & \bm{$79.0$} & \cellcolor{violet!15}{\bm{$79.4$}} & \bm{$75.4$} & \bm{$75.0$} & \bm{$68.8$} & \bm{$72.0$} & \bm{$64.4$} & \bm{$57.6$} \\
    \specialrule{1pt}{0.2\jot}{0.1pc}
    \cellcolor{blue!15} & Basic prompts & 52.52 & \cellcolor{violet!15}54.36 & 53.70 & 50.54 & 47.44 & 43.10 & 36.38 & 33.20 & 28.06 & \multirow{4}{*}{83.34}\\
    \cellcolor{blue!15} & CiP (zero-shot)& $51.88$ & \cellcolor{violet!15}$53.36$ & $52.64$ & \color{orange}$51.68$ & \color{orange}$49.18$ & \color{orange}$44.24$ & \color{orange}$41.56$ & \color{orange}$39.00$ & \color{orange}$34.00$ \\
     \cellcolor{blue!15} & CiP (ViT-GPT2) & $52.66$ & $56.38$ & \cellcolor{violet!15}$57.04$ & $56.66$ & $55.18$ & $52.00$ & $48.18$ & $46.58$ & $42.08$ & \\
     \multirow{-4}{*}{\cellcolor{blue!15}\rotatebox{90}{INet-100}} & CiP (BLIP-2) & \bm{$59.28$} & \bm{$61.56$} & \cellcolor{violet!15}\bm{$62.38$} & \bm{$61.64$} & \bm{$60.16$} & \bm{$55.68$} & \bm{$53.34$} & \bm{$47.36$} & \bm{$44.92$} & \\
    \bottomrule
    \end{tabular}
    }
    \label{tab:imagenette&100}
\end{table*}

In this section, we evaluate the performance of CiP on multiple popular image datasets from the aspect of generalization, OoD robustness, privacy preservation, and data augmentation. Except the main experiments, we also provide ablation studies of different architectures and settings and empirical analysis of the performance gain. 

\paragraph{Datasets} We conduct experiments on three image datasets of ImageNette \citep{imagewang}, ImageNet-100 \citep{tian2020contrastive}, and ImageNet-1K \citep{russakovsky2015imagenet}. ImageNet-1K is a large-scale image dataset with $1000$ classes and each class has around $1300$ images. 
ImageNette contains images of $10$ easy-to-classify classes selected from ImageNet-1K, while ImageNet-100 contains $100$ classes randomly sampled from ImageNet-1K.

\paragraph{Pre-trained Models} We adopt two models of ViT-GPT2 \citep{kumar2022imagecaptioning} and BLIP-2 \citep{li2023blip} to generate captions for target image dataset, and BLIP-2 captions are default used unless otherwise stated. The powerful Stable Diffusion (SD)\footnote{We adopt Stable Diffusion v1.5 from Hugging Face. \url{https://huggingface.co/runwayml/stable-diffusion-v1-5}} is used as the T2I model to generate synthetic images by feeding prompts, and the resolution of output images is $512\times512$. For SD, guidance scale (gs) is an important hyperparameter to control image generation. Concretely, a large guidance scale increases the consistency between the given textual prompt and the generated image, while a smaller guidance scale contributes to better generation diversity. 

\paragraph{Setup} To measure the training effect of generated synthetic dataset $\mathcal{S}$, we train ResNet-50 \citep{he2016deep} on $\mathcal{S}$ and then collect the classification accuracy {\it w.r.t.} the real data generating distribution $\mathcal{D}$. We also provide results on training Vision Transformer (ViT) \citep{dosovitskiy2020image} in ablation study. Details for training the model are presented in Appendix \ref{app:model training}. 

\subsection{Main Results}
\paragraph{ImageNette \& ImageNet-100}
For ImageNette and ImageNet-100, we generate corresponding synthetic datasets with different guidance scales of $[1, 1.5, 2, 2.5, 3, 4, 5, 6, 7.5]$ and different prompt strategies of (1) basic prompts of ``\texttt{a photo of \{class\}}'' as the baseline; (2) CiP with ViT-GPT2 generated captions; (3) CiP with BLIP-2 generated captions; (4) CiP with zero-shot setting (introduced in Section \ref{sec:ablation study}), and therefore total $9\times 4 = 36$ synthetic datasets are generated by Stable Diffusion for ImageNette and ImageNet-100, respectively. 
The Top-$1$ accuracy of these synthetic datasets is presented in Table \ref{tab:imagenette&100}. {We also present the accuracy of real datasets (Real data) for completion.} Notably, although \citet{sariyildiz2023fake} also measure the classification accuracy of models trained on synthetic ImageNet-100 generated by SD, we do not include their result in Table \ref{tab:imagenette&100} because their strategy of training ResNet-50 is different from ours, and a detailed discussion is presented in Appendix \ref{app:result comparison}. The following observations can be obtained from the table:
\begin{itemize}[leftmargin=5mm]
    \item \textbf{Guidance scale has significant influence.} Both large and small guidance scales decrease performance and thus are harmful to the training effect of synthetic datasets, and the proper range of gs is $[1.5, 2.0]$ for balancing the trade-off between generation diversity and consistency.
   
    \item \textbf{CiP remarkably improves training effect.} As shown in the results of ViT-GPT2 and BLIP-2, CiP increases classification accuracy by a substantial margin (around $10\%$) compared to basic prompts. This remarkable improvement indicates that our data-corresponded prompting diversifies synthetic images in a more controllable manner. We leave a detailed analysis in the next section.

    \item \textbf{Captioning model matters.} By comparing the results between different captioning models of ViT-GPT2 and BLIP-2, prompts based on BLIP-2 consistently achieve higher performance compared to the ViT-GPT2 model under different guidance scales. This suggests that CiP can further increase the training effect of generated synthetic datasets with a better captioning model. Therefore, it is a promising direction to design decent captioning models to facilitate training set generation with T2I models. In addition, from the reverse thinking, CiP can be equipped with different captioning models to evaluate their capability of captioning images.

\end{itemize}

\begin{table*}[t]
\scriptsize
    \centering
    \renewcommand{\arraystretch}{0.7}
    \caption{Top-$1$ and Top-$5$ accuracy ($\%$) of ResNet-50 on multiple ImageNet-1K variants. \textbf{Bold} scores denote the best results of synthetic data for each validation set, and {\color{orange} orange-colored} scores denote the performance higher than real data. INet-R$^\ast$ and INet-A$^\ast$ only cover $200$ classes of ImageNet-1K and our reported performances are evaluated on these $200$ classes.}
    \resizebox{\linewidth}{!}{
    \begin{tabular}{ccccc ccccc c}
    \toprule
     \multirow{2}{*}{Method}  & \multicolumn{2}{c}{INet-Val} & \multicolumn{2}{c}{INet-v2} & \multicolumn{2}{c}{INet-Sketch} & \multicolumn{2}{c}{INet-R$^\ast$}& \multicolumn{2}{c}{INet-A$^\ast$}    \\
     & Top-1 & Top-5 & Top-1 & Top-5 & Top-1 & Top-5 & Top-1 & Top-5 & Top-1 & Top-5\\
     \specialrule{1pt}{0.2\jot}{0.2pc}
     Real data & $79.56$ & $94.61$ & $74.71$ &  $92.20$ & $28.10$ & $45.77$ & $39.38$ & $54.10$ & $8.05 $ & $34.65$  \\
        \specialrule{1pt}{0.2\jot}{0.2pc}
    BigGAN \citep{ravuri2019seeing} & $42.65$  & $65.92$ & - & - & - & - & - & - & - & -  \\
     INet-SD \citep{sariyildiz2023fake} & $42.89$ & $70.26$ & $42.98$ &  $70.32$ & $16.59$ & $35.18$ & $26.29$ & $45.31$ & $3.55$ & $15.08$  \\
     Basic prompts & $45.23$ & $69.88$  & $45.64$ & $70.96$ & $17.68$ & $34.08$ & $30.12$ & $46.81$ & $4.17$ & $14.71$  \\
     LLM-Gen &  {$43.13$} &  {$68.03$}  &  {$43.42$} &  {$68.58$} &  {$14.29$} &  {$28.76$} &  {$25.67$} &  {$41.79$} &  {$5.37$} &  {$18.76$} \\
      {LLM-Sum} \citep{dunlap2023diversify} &  {$53.27$} &  {$79.83$}  &  {$52.83$} &  {$79.99$} &  {$15.92$} &  {$30.95$} &  {$31.40$} &  {$48.28$} &  {$5.79$} &  {$22.96$}  \\
     
     \specialrule{1pt}{0.2\jot}{0.2pc}
     {CiP} & {$54.06$} & {$80.51$} & {$53.78$} & {$80.47$} & {$18.47$} & {$35.47$} & {$33.57$} & {$51.06$} & {$5.19$} & {$21.68$}  \\
     {CiP} ($5\times$) & $60.95$ & $86.26$ & $60.30$ &  $86.08$ & $24.73$ & $45.30$ & \color{orange}$42.94$ & \color{orange}$61.91$ & \color{orange}$9.17 $ & $32.45$  \\
     {CiP} ($10\times$) & \bm{$62.14$} & \bm{$87.07$} & \bm{$61.40$} & \bm{$86.84$} & \bm{$26.60$} & \color{orange}\bm{$47.86$} & \color{orange}\bm{$45.08$} & \color{orange}\bm{$64.71$} & \color{orange}\bm{$10.44$} & \bm{$34.44$}  \\
    \bottomrule
    \end{tabular}
    }
    \label{tab:imagenet1k}
\end{table*}

\paragraph{ImageNet-1K} For the large-scale ImageNet-1K, we generate corresponding synthetic datasets with the gs of $1.5$ and different strategies of basic prompts and CiP, respectively.  {For a comprehensive comparison, we construct the prompt set by utilizing the powerful LLM of Vicuna-13B \citep{vicuna2023} to: (1) generate $13$ image captions and for each class (LLM-Gen) and (2) follow \citep{dunlap2023diversify} to summarize $130$ image captions from approximately $1300$ real image captions per class (LLM-Sum).} We compare to the ImageNet-1K-like synthetic data generated by BigGAN \citep{ravuri2019seeing}, SD v1.4 (ImageNet-SD) \citep{sariyildiz2023fake}, {LLM-Gen, and LLM-Sum} in Table \ref{tab:imagenet1k}. By evaluating well-trained models on ImageNet validation set (INet-Val) and ImageNet-v2 (INet-v2) \citep{recht2019imagenet}, we observe that (1) synthetic data generated by SD v1.5 ($\text{gs}=1.5$) with basic prompts have a similar performance to those generated by SD v1.4 ($\text{gs}=2$) with prompts of \texttt{``\{class\}, \{definition of the class\}''}; (2) CiP substantially increases the performance in ImageNet validation sets compared to Basic prompts ($45.23\%\rightarrow 54.06\%$ on INet-Val Top-1), suggesting that synthetic data induced by CiP better characterize the original data distribution;  {(3) LLM-Gen performs slightly worse than Basic prompts, which we attribute to the subjective imagination of LLM. A detailed analysis {\it w.r.t.} LLM-generated/polished captions is provided in Section \ref{sec:ablation study}. (4) CiP outperforms LLM-Sum, underscoring the critical importance of a diverse prompt set in training data synthesis;} and (5) by generating $10\times$ synthetic data {\it w.r.t.} ImageNet-1K via CiP, corresponding well-trained models have a large performance gain ($54.06\%\rightarrow 62.14\%$ on INet-Val Top-1). This remarkable improvement on training effect of synthetic data suggests that {\it class-relevant textual information in prompts can successfully transfer to corresponding visual information via T2I models}.

\paragraph{Out-of-distribution Robustness} Other than testing on in-distribution data, we further measure the performance on multiple challenging ImageNet variants including ImageNet-Sketch (INet-Sketch) \citep{wang2019learning}, ImageNet-R (INet-R) \citep{hendrycks2021many}, and ImageNet-A (INet-A) \citep{hendrycks2021natural} which contains different types of out-of-distribution images. The results in Table \ref{tab:imagenet1k} show that (1) all synthetic datasets with the same size of ImageNet-1K have inferior OoD performance to real data, while CiP achieves the best among them; and (2) when more synthetic data generated with CiP, the OoD performance is more efficiently increased compared to in-distribution performance and finally {\it surpasses real data}, which is firstly achieved in ImageNet-1K by T2I synthetic data. Therefore, CiP can synthesize fairly informative OoD examples that enhance models' robustness and contribute to their safe deployment in the open world.

\begin{figure}[t]
\begin{minipage}{0.41\textwidth}
\centering
\captionof{table}{{MIA accuracy {\it w.r.t.} ResNet-50 trained on real and synthetic data.}}
\resizebox{\linewidth}{!}{
    \begin{tabular}{ccccc}
    \toprule
    & Real & Basic & CiP\\
    \midrule
        MIA Acc & $0.58$  & $0.50$ & $0.50$   \\
    \bottomrule
    \end{tabular}
    }
\label{tab:mia}
\vspace{-0.15cm}
\end{minipage}
\hfill
\begin{minipage}{0.49\textwidth}
\captionof{table}{{Top-1 accuracy ($\%$) of ResNet-50 trained by real data and combined (real+synthetic) data.}}
\centering

\resizebox{\linewidth}{!}{
    \begin{tabular}{cccc}
    \toprule
    & Real & Real + Basic & Real + CiP\\
    \midrule
    INette & $91.4$ & $92.8({\color{ForestGreen} 1.4 \uparrow})$& \bm{$94.4$}$({\color{ForestGreen} 3.0 \uparrow})$ \\
    INet-100 & $83.34$ & $84.22 ({\color{ForestGreen} 0.88 \uparrow})$  & \bm{$84.52$} $({\color{ForestGreen} 1.18 \uparrow})$ \\
    \bottomrule
    \end{tabular}
    }
\label{tab:data aug}
\end{minipage}
\end{figure}

\paragraph{Privacy Analysis} We evaluate the capability of privacy preservation of our method via membership inference attack (MIA), which infers whether a given data example comes from the target model's training dataset \citep{shokri2017membership,yeom2018privacy}; please see Appendix \ref{app:mia} for detail computation. As shown in Table \ref{tab:mia}, The MIA accuracy on models trained on real data is $0.58$, {\it i.e.}, attackers have a success rate of $58\%$ to infer whether a given example is from the training set or the test set. As Basic Prompt does not contain any prior knowledge about real training data, it routinely achieves the best MIA {accuracy} of $0.5$. Although our approach collects image captions from real training data, the MIA {accuracy} {\it w.r.t.} CiP is still $0.5$, {\it i.e.}, random guess, which suggests that {\it CiP completely makes downstream models immune to membership inference attacks} and thus prevents information leakage from real data. 

\paragraph{CiP as Data Augmentation} We further evaluate our CiP in data augmentation setting, {\it i.e.}, the synthetic data are trained with real data together. Previous study shows that synthetic data generated by fine-tuned generative models has marginal improvement when trained with real data \citep{ravuri2019classification,bansal2023leaving}. This might be attributed to that the generative model memorizes the real data during fine-tuning. As the T2I model does not directly touch the real data but is solely fed with prompts in our CiP, it is expected to generate more diverse images compared to real data and achieve better augmentation performance. We conduct the data augmentation experiments by training ResNet-50 on combined data composed of the same number of real and synthetic data; the results are presented in Table \ref{tab:data aug}. From this table we observe that synthetic data generated with CiP can improve ImageNette by $3.0\%$ and ImageNet-100 by $1.18\%$, respectively. These improvements imply that CiP is a reliable method to augment the real data for further enhancing the model generalization.

\subsection{Ablation Study}
\label{sec:ablation study}
\paragraph{Multi-architecture Experiments} In addition to obtaining performance on ResNet-50, we measure the training effect of synthetic data on the popular Vision Transformer architecture ViT-B/16 \citep{dosovitskiy2020image}. Note that all ViTs are trained on real or synthetic ImageNet-1K from scratch, and the results are shown in Table \ref{tab:vit}, from which we observe that CiP consistently increases the performance of generated synthetic datasets on multiple ImageNet variants by a large margin compared to basic prompts. Therefore, the improvement of the training effect brought by CiP is general and significant {\it w.r.t.} different network architectures.

\paragraph{Zero-shot CiP} Except for conducting CiP on real images to produce the high-quality prompt set, we also investigate zero-shot CiP in which the real data are not given for captioning. In this setting, we first generate synthetic images by feeding the basic prompt of ``\texttt{A photo of \{class\}}'' to T2I model and collect all \textit{preliminary synthetic images} $\mathcal{S}^\prime$. Then we construct the prompt set by captioning these preliminary synthetic images $\mathcal{S}^\prime$ and generate the final synthetic dataset $\mathcal{S}$. By comparing the result of basic prompts and zero-shot CiP in Table \ref{tab:imagenette&100}, we observe that although zero-shot CiP boosts the performance of basic prompts when gs is large (gs $\geq 4$), it hardly improve the performance compared to the basic prompt under a proper gs. This illustrates the importance of incorporating the class-relevant information into textual prompts, and naively augmenting prompts without any prior knowledge might be insignificant to induce informative synthetic data.

\begin{table*}[t]
\scriptsize
    \centering
    \renewcommand{\arraystretch}{0.7}
    \caption{Top-$1$ and Top-$5$ accuracy ($\%$) of ViT-B/16 on ImageNet-1K variants. \textbf{Bold} scores denote the best results of synthetic data for each validation set.  INet-R$^\ast$ and INet-A$^\ast$ only cover $200$ classes of ImageNet-1K and our reported results are evaluated on these $200$ classes.}
    \resizebox{\linewidth}{!}{
    \begin{tabular}{ccccc ccccc c}
    \toprule
     \multirow{2}{*}{Method}  & \multicolumn{2}{c}{INet-Val} & \multicolumn{2}{c}{INet-v2} & \multicolumn{2}{c}{INet-Sketch} & \multicolumn{2}{c}{INet-R$^\ast$}& \multicolumn{2}{c}{INet-A$^\ast$}    \\
     & Top-1 & Top-5 & Top-1 & Top-5 & Top-1 & Top-5 & Top-1 & Top-5 & Top-1 & Top-5\\
     \specialrule{1pt}{0.3\jot}{0.2pc}
     Real data & $78.24$  &  $93.36$ & $72.93$ & $90.37$ & $26.01$ & $42.98$ & $36.84$ & $51.77$ & $15.12 $ & $43.01$  \\
        \specialrule{1pt}{0.2\jot}{0.2pc}
    Basic prompts & $44.00$  & $70.30$ & $44.76$ & $70.64$ & $17.92$ & $35.00$ & $31.45$ & $50.23$ & $4.77$ & $16.72$  \\
     {CiP} & \bm{$53.04$} & \bm{$79.31$} & \bm{$52.61$} & \bm{$79.49$} & \bm{$18.10$} & \bm{$35.39$} & \bm{$34.20$} & \bm{$51.86$} & \bm{$6.92$} & \bm{$25.51$}  \\

    \bottomrule
    \end{tabular}
    }
    \label{tab:vit}
\end{table*}

\begin{table*}[t]
    \centering
    \caption{Top-1 accuracy ($\%$) of ResNet-50 trained on synthetic ImageNette. {\color{red} Red-colored scores} in brackets denote the performance drop compared to original captions without rewriting.}
    \resizebox{\linewidth}{!}{
    \begin{tabular}{ccccc ccccc}
    \toprule
     \multirow{2}{*}{Methods} & \multicolumn{9}{c}{Guidance Scale}  \\
    \cmidrule{2-10}
    & $1$ & $1.5$ & $2$ & $2.5$ & $3$ & $4$ & $5$ & $6$ & $7.5$  \\
    \specialrule{1pt}{0.2\jot}{0.2pc}
      CiP (zero-shot) + LLM & \format{56.0}{9.8} & \format{57.0}{10.0} & \format{58.6}{7.0} & \format{55.8}{10.6} & \format{59.6}{2.4} & \format{52.8}{8.8} & \format{45.8}{10.2}  & \format{48.4}{7.4} & \format{42.2}{7.9} \\
    CiP (BLIP-2) + LLM & \format{67.8}{9.6} & \format{66.8}{12.2}  & \format{67.6}{11.8}  & \format{64.6}{10.8} & \format{60.8}{14.2} & \format{63.2}{5.6}   & \format{57.0}{15.0} & \format{55.0}{9.4}&\format{45.4}{12.2}  \\
    \bottomrule
    \end{tabular}
    }
    \label{tab:imagenette llm}
\end{table*}

\begin{figure}
  \centering
  \begin{subfigure}{0.48\linewidth}
    \includegraphics[width=0.9\columnwidth]{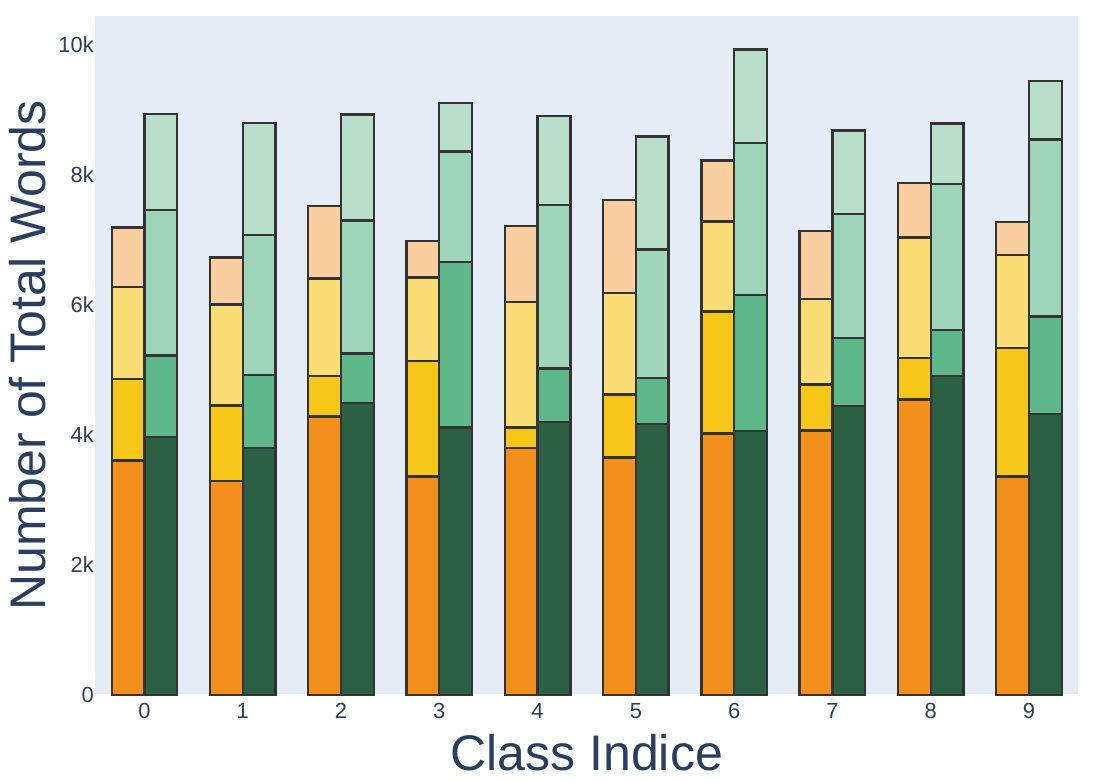}
    \caption{number of total words}
    \label{fig:total words}
  \end{subfigure}
  \hfill
  \begin{subfigure}{0.48\linewidth}
    \includegraphics[width=0.9\columnwidth]{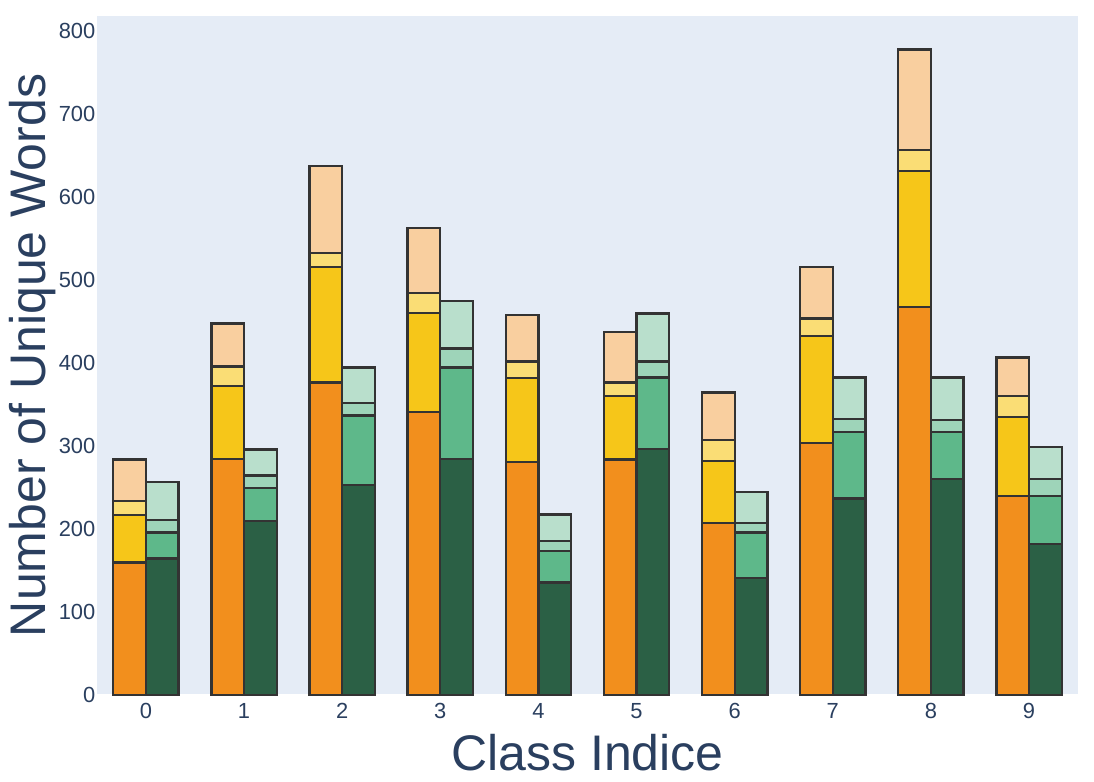}
    \caption{number of unique words}
    \label{fig:unique words}
  \end{subfigure}
  \caption{The number of (a) total words and (b) unique words in captions produced by BLIP-2 and ViT-GPT2 on each class of ImageNette. Multiple color bars are used to denote the different parts of speech of in captions. For BLIP-2: {\color{blip_noun}\small $\blacksquare$} (noun)  {\color{blip_verb}\small $\blacksquare$} (verb) {\color{blip_prep}\small $\blacksquare$} (prep.)  {\color{blip_adj}\small $\blacksquare$} (adj.), and for ViT-GPT2: {\color{vit_noun}\small $\blacksquare$} (noun)  {\color{vit_verb}\small $\blacksquare$} (verb) {\color{vit_prep}\small $\blacksquare$} (prep.) {\color{vit_adj}\small $\blacksquare$} (adj.).}
  \label{fig:word stat}
\end{figure}

\paragraph{Different Captioning Models} As shown in Table \ref{tab:imagenette&100}, BLIP-2 shows a superior performance to ViT-GPT2, and we investigate this by analyzing the informative words (noun, verb, preposition, adjective) in their captions. We plot the total number of informative words and the number of unique words {\it w.r.t.} their captions in Figure \ref{fig:word stat}. From the plots we observe that although ViT-GPT2 generates longer captions (Fig. \ref{fig:total words}), BLIP-2 can provide more unique informative words across different classes (Fig. \ref{fig:unique words}), which contribute to more diverse behaviors and backgrounds in synthetic data.

\paragraph{Polishing Captions with LLM} As LLMs have shown unprecedented performance in multiple NLP tasks, a natural idea is to polish T2I prompts via LLMs. Here we use Vicuna-13B \citep{vicuna2023} to polish captions generated by BLIP-2 with the following prompt:
\begin{tcolorbox}
[notitle,boxrule=0pt,colback=blue!10,colframe=blue!10]
\texttt{\small This is an image caption about \{class name\} category. Can you unemotionally and succinctly rewrite it to 2 captions by containing the word of \{class name\} in more diverse scenarios? \\
\# Caption: \\ 
\{caption\} \\
\# Answer:
}
\end{tcolorbox}
The results are presented in Table \ref{tab:imagenette llm}, and we observe that there is a large performance drop (red-colored numbers, about $10\%$) after the LLM polish. We show some rewritten examples in Tables \ref{tab:llm caption blip2} and \ref{tab:llm caption zero-shot} in Appendix \ref{app:llm rewrited caption}. By analyzing the generated captions by Vicuna, we found that the LLM focuses on (1) supplementing characteristics of the class name and (2) adding subjective adverbs ({\it e.g.}, proudly, eagerly, casually) and imagination ({\it e.g.}, celebrating a memorable fishing trip, prepares for a tournament), which are hardly shown by image data and also distract the T2I model to ignore class objects, thereby causing the notable drop.

\subsection{Empirical Analysis}

In this section, we study the performance gain of CiP by in-depth investigating the generated captions and images. We observe that the main improvement can be attributed to:

\renewcommand\labelenumi{(\theenumi)}
\begin{enumerate}[leftmargin=7mm]
    \item \textbf{Class-relevant information extraction.} The image captions exactly extract class-relevant information, including backgrounds and behaviors, from real data. For better understanding, we present two captions as follows and highlight the corresponding backgrounds and behaviors in green and red boxes, respectively.
    \begin{center}
        \texttt{\small English Springer: ``two dogs \fcolorbox{white}{red!15}{standing} next to a tree \fcolorbox{white}{green!15}{in the snow}'' \\
        tench: ``a large yellow fish \fcolorbox{white}{red!15}{swimming} \fcolorbox{white}{green!15}{in an aquarium}'' \\
        }
    \end{center}
    As demonstrated in the above captions, \texttt{English Springer} and \texttt{tench} are in different backgrounds (snow {\it vs.} aquarium) and show different behaviors (stand {\it vs.} swim). According to Table \ref{tab:real_blip}, the core information of real data is precisely captured by captions and well restored by the synthetic images. In contrast, the synthetic images generated by basic prompts often show the object in a monotonous manner and \textit{lack interactions between different objects or between objects and backgrounds}. Consequently, such class-relevant augmentation produced by CiP is more reasonable than augmenting data with class-agnostic backgrounds, which has been observed to marginally increase the training effect of synthetic data \citep{sariyildiz2023fake}.

    \item \textbf{Clarification of polysemy}. Some words of class names possess multiple meanings, which impedes the T2I models to understand given class names and consequently induces unfaithful images. We provide captions as below and highlight the main object with the blue box, which helps correct the misreading by T2I models.
    \begin{center}
    \texttt{\small jay: ``a \fcolorbox{white}{blue!15}{bird} perched on a metal bird feeder'' \\
    baseball: ``a box of \fcolorbox{white}{blue!15}{baseballs} in plastic wrap'' \\
    baseball: ``a baseball \fcolorbox{white}{blue!15}{player} swinging a bat at a ball''}
    \end{center}
    For instance, the class label \texttt{jay}, representing a bird species in the dataset, is frequently interpreted as a person's name by Stable Diffusion in the absence of additional context or instruction. This misunderstanding can be largely alleviated by providing additional annotations, {\it e.g.}, ``a bird perched on a metal bird feeder'' in the prompt. Apart from this complete misunderstanding, there is some more tricky polysemy that the class name can represent multiple similar concepts. As shown in the last row of Table \ref{tab:real_blip}, the class name of \texttt{baseball} is often considered as a kind of activity by the T2I model without additional annotation and thus images with baseball fields and audience are synthesized. In contrast, the captions help clarify main objects in images and consequently induce faithful synthetic images.
\end{enumerate}

According to the above analysis, a good prompt set is expected to (1) possess diverse behaviour and background descriptions; (2) be succinct and objective so that the T2I model can faithfully extract class-relevant information from the prompts with less hallucination, which shed light on future prompt design {\it w.r.t.} training effect of synthetic data.

%% file: sec/conclusion.tex
\section{Discussion and Conclusion}

\paragraph{Discussion} 
Although our experiments utilize captions extracted from real images to explore the performance limits of CiP, {these captions can alternatively be sourced from existing datasets such as COCO \citep{lin2014microsoft} and Visual Genome \citep{krishna2017visual}}, as well as diverse text-based resources relevant to the target domain, including books, online content, and publicly available language datasets. This highlights the fact that real-world image data is not a prerequisite for the application of CiP.

With regard to data augmentation methods, their primary objective is to generate variations of real images through transformations such as rotation, cropping, and flipping. However, these techniques inherently depend on real images as input, and the resulting augmented images often retain significant similarity to the originals, thereby posing a potential risk of data leakage. In contrast, CiP relies solely on text prompts as input to T2I models, producing synthetic images that are inherently distinct from real images and effectively mitigating any risk of data leakage, even when captions are extracted from real images (see Table \ref{tab:mia}). Furthermore, the diverse set of images generated by CiP is fully compatible with conventional data augmentation techniques, allowing for additional enhancements and further improvements in downstream tasks (see Table \ref{tab:data aug}). Consequently, CiP is not merely a data augmentation technique, which use real images as inputs, but rather an effective solution for safe and large-scale training with purely synthetic data.

\paragraph{Generation efficiency} Albeit promising on privacy preservation and other applications, training data synthesis via large diffusion models requires much computing resources. For example, $350$ GPU hours (with NVIDIA V100) are needed to generate one ImageNet-1K-like dataset. Reducing the synthesis cost is important for paving the way for large-scale and edge-computing applications.

\paragraph{Conclusion} In this paper we investigate the influence of prompt design on the training effect of synthetic data. We propose Caption in Prompt to induce highly-informative synthetic data from large T2I models for downstream model training. The main idea is to generate high-quality prompts using the off-the-shelf image captioning models. Empirical study shows that CiP facilities training set synthesis by extracting class-relevant information and clarifying polysemy, thereby contributing to more rational and diverse synthetic data. Extensive experiments on large-scale image datasets show that CiP significantly improves the model generalization, OoD robustness, and privacy protection and is also an effective data augmentation strategy.

%% file: sec/supplementary.tex








\clearpage
\appendix

\section{Implementation Details}
\label{app:implementation}
This section provides all the additional implementation details of our experiments.

\subsection{Downstream Model Training.}
\label{app:model training}
\paragraph{ResNet-50} During the training procedure of ImageNette and ImageNet-100, we employ SGD to optimize ResNet-50 for $200$ epochs and the momentum factor is $0.9$. The weight decay factor is set to $5$e-$4$, and the initial learning rate is $0.1$ and decayed by $0.2$ every $50$ epochs. Besides, basic data augmentation (crop and flip) \citep{Zagoruyko2016WRN} is adopted in the training. ImageNette is trained on single NVIDIA V100 GPU with batch size $128$, and Imagenet-100 is trained on 4 NVIDIA V100 GPUs with batch size $512$. For ImageNet-1K, we mainly follow the training recipe in \citep{wightman2021resnet} on 4 NVIDIA A100 GPUs with a total batch size of $2,048$. We adopt Lamb optimizer \cite{you2019large} with a initial learning rate of $0.005$ and a weight decay of $0.01$. We use cosine scheduler to anneal the learning rate with total 300 epochs and the first 5 epochs as warm-up. We utilize RandAugment \cite{cubuk2020randaugment}, Mixup \cite{zhang2017mixup} of 0.2, and CutMix \cite{yun2019cutmix} of 1.0 for data augmentation. The Network is trained using Binary Cross-Entropy (BCE) loss, with a smoothing of $0.1$.

\paragraph{ViT} For Vision Transformer (ViT) \cite{dosovitskiy2020image}, we train it only on ImageNet-1K, following the training recipe in \cite{touvron2022deit}. A Lamb optimizer with initial learning rate of $0.003$ and a weight decay of $0.02$. Similarly, the learning rate is cosine-annealed for $300$ epochs with $5$ warm-up epochs. The model is trained on 8 NVIDIA A100 GPUs with a total batch size of $2,048$ with BCE loss (no smoothing). We utilize RandAugment \cite{cubuk2020randaugment}, Mixup \cite{zhang2017mixup} of 0.8, and CutMix \cite{yun2019cutmix} of 1.0 as data augmentation. We additionally adopt color jitter of $0.3$ in data augmentation.

\subsection{Additional Details in Zero-shot CiP}
For zero-shot CiP, we first generate the preliminary synthetic dataset via SD with guidance scale of $1.5$. Then CiP is applied to the preliminary synthetic dataset.

\subsection{Vicuna-13B Generation}

We utilize Vicuna-13B \citep{vicuna2023} to (1) directly generate captions; (2) summarize the generated captions; and (3) rewrite the generated captions. We set the max tokens of Vicuna-13B as $512$, and temperature as $0.7$.

{
\paragraph{Prompt for Generating Captions}
    \begin{tcolorbox}
[notitle,boxrule=0pt,colback=blue!10,colframe=blue!10]
\texttt{\small For category of \{class name\}, please generate 15 unique image captions with different objective descriptions and backgrounds. \\
Image captions should be started with ``a photo of \{class name\}''.\\
The output captions should be:
}
\end{tcolorbox}

\paragraph{Prompt for Summarizing Captions}
\begin{tcolorbox}
[notitle,boxrule=0pt,colback=blue!10,colframe=blue!10]
\texttt{\small I have a set of image captions about \{class name\} category that I want to summarize into objective descriptions that describe the scenes, actions, camera pose, zoom, and other image qualities present. \\
My captions are: \\
\{caption\}\\
I want the output to be 130 summarized captions. \\
Each summarized caption describes a unique setting, and should be started with {prefix} to contain \{class name\}.\\
Based on the above captions, the output should be:
}
\end{tcolorbox}
}

\paragraph{Processing of Rewritten Texts} After obtaining the raw output, we use regular expression to process them. We extract the raw text following \textit{\# Answer}, remove all the emoji, and maintain only alphabetic characters. We random select $1$ caption from the generated $5$ caption candidates to maintain roughly the same size as the original captions.

\subsection{Computing MIA Accuracy}
\label{app:mia}
We utilize a threshold-based version of MIA \citep{yeom2018privacy} to evaluate our method in protecting privacy. Given the dataset $\mathcal{S}=(\mathcal{S}_{\text{train}}, \mathcal{S}_{\text{test}})$, where $\mathcal{S}_{\text{train}}$ and $\mathcal{S}_{\text{test}}$ are real training set and real test set, respectively.

Suppose the ``to-be-inferenced'' example $(\mathbf{x},y)$ comes from $\mathcal{S}$, then the accuracy of membership inference attack with a threshold $\zeta$ is calculated as follow,
\begin{equation}
\begin{aligned}
\operatorname{Acc}(\zeta) = \frac{1}{2} \times \left( \frac{\sum_{(\mathbf{x},y) \in \mathcal{S}_{\text{train}}} \bm{1}[\hat{p}\left(\mathbf{x}\right) \geq \zeta]}{\vert \mathcal{S}_{\text{train}} \vert}  + \frac{\sum_{(x,y) \in \mathcal{S}_{\text{test}}} \bm{1}[\hat{p}\left(\mathbf{x}\right) < \zeta]}{\vert \mathcal{S}_{\text{test}} \vert} \right).
\end{aligned}
\end{equation}
Then, we can find the optimal threshold $\zeta_{\text{optim}}$ with maximizing the attack accuracy, {\it i.e.},
\begin{equation}
\zeta_{\text{optim}} = \arg \max_{\zeta} \operatorname{Acc}(\zeta),
\end{equation}
and $\operatorname{Acc}(\zeta_{\text{optim}})$ is the final attack accuracy for $f_{\boldsymbol{\theta}}$. Lower $\operatorname{Acc}(\zeta_{\text{optim}})$ means the attacker can hardly judge whether an example $\mathbf{x}\in \mathcal{S}$ belongs to $S_{\text{train}}$ or $S_{\text{test}}$, {\it i.e.}, the attacker infers less information from the model.

\section{Proof of Theorem \ref{thm:1}}
\label{app:proof of thm 1}
\begin{proof} 
Recall the definition of expected risk $\mathcal{R}_{\mathcal{D}}$, we have
\begin{align}
     \mathcal{R}_{\mathcal{D}}&(\texttt{alg}(\mathcal{S})) - \mathcal{R}_{\mathcal{D}_{\mathcal{C}}}(\texttt{alg}(\mathcal{S})) \\
     =& \mathbb{E}_{(\boldsymbol{x},y) \sim \mathcal{D}}\left[\mathbb{I} \left( \mathcal{M}_{\texttt{alg}(\mathcal{S})}\left( \boldsymbol{x} \right) \neq y  \right) \right] \\
     &- \mathbb{E}_{(\boldsymbol{x},y) \sim \mathcal{D}_{\mathcal{C}}}\left[\mathbb{I} \left( \mathcal{M}_{\texttt{alg}(\mathcal{S})}\left( \boldsymbol{x} \right) \neq y  \right) \right] \nonumber \\
     =& \int \left( \mathbf{P}_{\mathcal{D}}(\boldsymbol{x}) \mathbf{P}_{\mathcal{D}}(y | \boldsymbol{x}) - \mathbf{P}_{\mathcal{D}_\mathcal{C}}(\boldsymbol{x}) \mathbf{P}_{\mathcal{D}_\mathcal{C}}(y | \boldsymbol{x}) \right) \\
     &\cdot \mathbb{I}  \left( \mathcal{M}_{\texttt{alg}(\mathcal{S})}\left( \boldsymbol{x} \right) \neq y  \right) \mathbf{d}\boldsymbol{x} \mathbf{d} y \\
     \leq& \int  \left\vert( \mathbf{P}_{\mathcal{D}}(\boldsymbol{x}) \mathbf{P}_{\mathcal{D}}(y | \boldsymbol{x}) - \mathbf{P}_{\mathcal{D}_\mathcal{C}}(\boldsymbol{x}) \mathbf{P}_{\mathcal{D}_\mathcal{C}}(y | \boldsymbol{x}) \right\vert  \mathbf{d}\boldsymbol{x} \mathbf{d} y \\
     =& \int \mathbf{P}_{\mathcal{D}}(y | \boldsymbol{x}) \left\vert( \mathbf{P}_{\mathcal{D}}(\boldsymbol{x}) - \mathbf{P}_{\mathcal{D}_\mathcal{C}}(\boldsymbol{x}) \right\vert  \mathbf{d}\boldsymbol{x} \mathbf{d} y \\
     =& \int \left\vert( \mathbf{P}_{\mathcal{D}}(\boldsymbol{x}) - \mathbf{P}_{\mathcal{D}_\mathcal{C}}(\boldsymbol{x}) \right\vert  \mathbf{d}\boldsymbol{x}
\end{align}
The proof is completed.
\end{proof}

\section{Comparison to \citep{sariyildiz2023fake}}
\label{app:result comparison}
In the ImageNet-100 experiments of \citep{sariyildiz2023fake}, the authors train the downstream ResNet-50 with cosine annealing learning rate and the powerful data augmentation of DINO \citep{caron2021emerging}, while monotonous learning decay and basic augmentation are adopted in our ImageNet-100 experiments. These differences in the training stage causes large accuracy gaps when training with real and synthetic data, and we thus do not include the ImageNet-100 results of \citep{sariyildiz2023fake} in our experiments. We present the results {\it w.r.t.} real data in Table \ref{tab:real data}. As for ImageNet-1K, the training strategy and also the classification accuracy of real data of \citep{sariyildiz2023fake} and ours are similar, and their results of ImageNet-1K are comprised in Table \ref{tab:imagenet1k} for a fair comparison.

\begin{table}[h]
    \centering
    \caption{Top-1 ACC ($\%$) of ResNet-50 with the training settings of  
 \citep{sariyildiz2023fake} and ours.}
    \begin{tabular}{c|c|c}
    \toprule
         &  ImageNet-100 & ImageNet-1K\\
    \midrule
    ImageNet-SD \citep{sariyildiz2023fake} & $87.4$ & $80.1$ \\
    Our basic prompts & $83.3$ & $79.5$ \\
    \bottomrule
    \end{tabular}
    \label{tab:real data}
\end{table}

\section{Examples of Syn Images and Captions}

\subsection{Synthetic Images Generated by Different Captions.} 
We show the captions generated by different models of ViT-GPT2 and BLIP-2 and captioned on real data or preliminary synthetic data (zero-shot setting) and the corresponding generated images with $\texttt{guidance scale}=2$, as shown in Table \ref{tab:caption and images}.

\begin{table*}[t]
    \centering
    \caption{Synthetic Images Generated by ViT-GPT2, BLIP-2, and zero-shot BLIP-2, respectively.}
    \resizebox{\linewidth}{!}{
    \begin{tabular}{c|p{3cm}c|p{3cm}c|p{3cm}c}
    \toprule
    & \multicolumn{2}{c|}{CiP (ViT-GPT2)} &  \multicolumn{2}{c|}{CiP (BLIP-2)} & \multicolumn{2}{c}{CiP (zero-shot)}  \\
    \midrule
      \multirow{12}{*}{\rotatebox{90}{Tench}} &  a fish in the middle of a grassy area. &  \raisebox{-0.9\totalheight}{\includegraphics[width=0.1\textwidth]{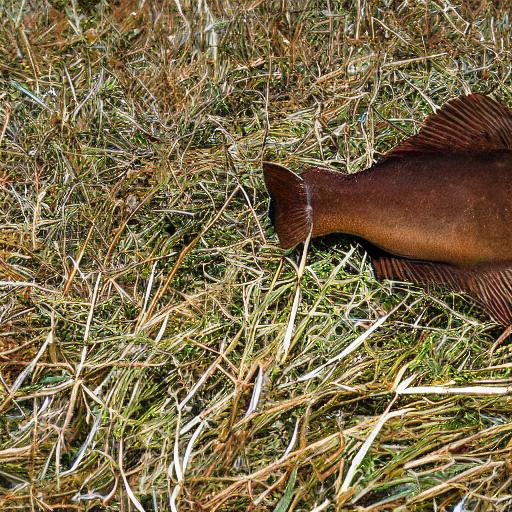}} & a fish laying on the grass in the grass. & \raisebox{-0.9\totalheight}{\includegraphics[width=0.1\textwidth]{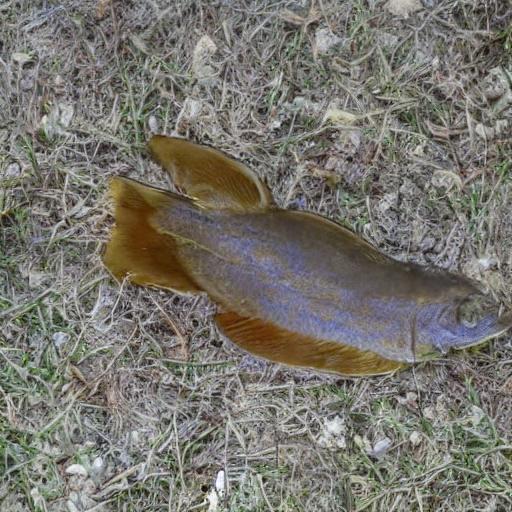}} & a bench in a wet area with trees and bushes & \raisebox{-0.9\totalheight}{\includegraphics[width=0.1\textwidth]{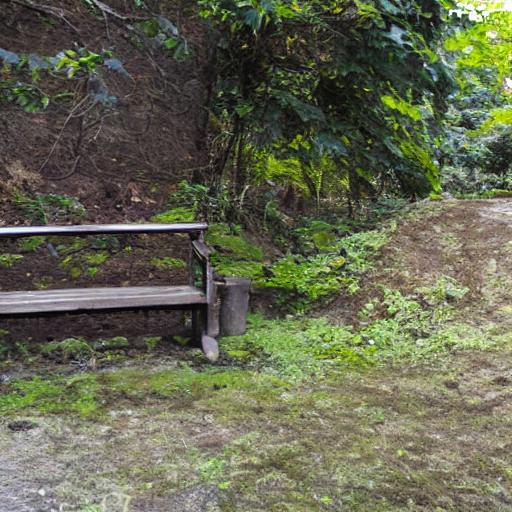}}\\
        & a man sitting on top of a green plant in a field. &  \raisebox{-0.9\totalheight}{\includegraphics[width=0.1\textwidth]{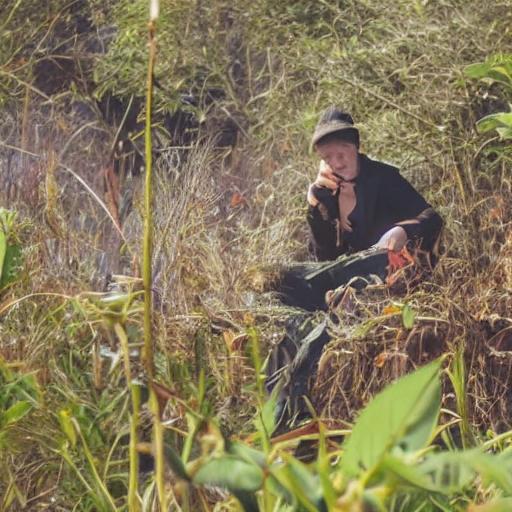}} & a man kneeling down holding a large fish. & \raisebox{-0.9\totalheight}{\includegraphics[width=0.1\textwidth]{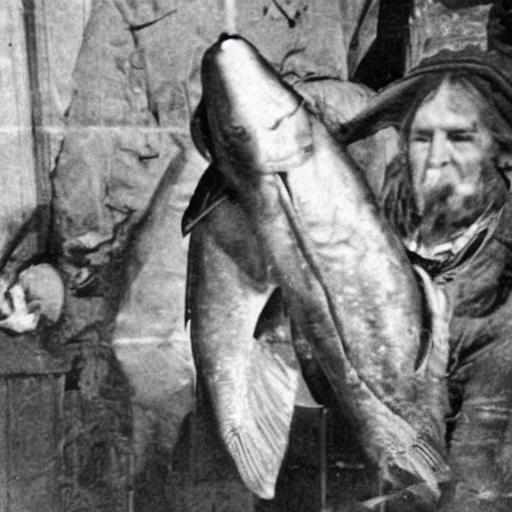}} & a large fish laying on some rocks & \raisebox{-0.9\totalheight}{\includegraphics[width=0.1\textwidth]{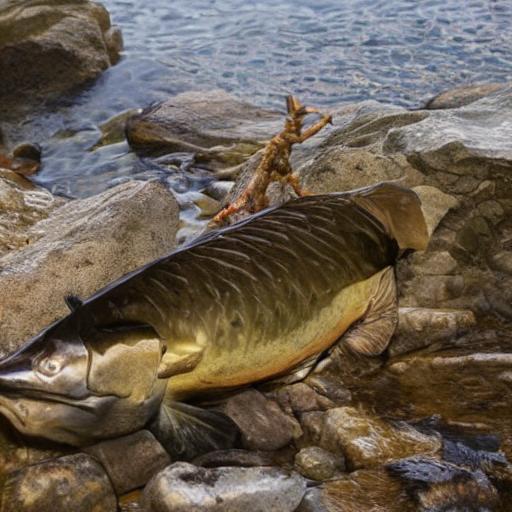}}\\
    & a man sitting on the side of a river holding a stick. &  \raisebox{-0.9\totalheight}{\includegraphics[width=0.1\textwidth]{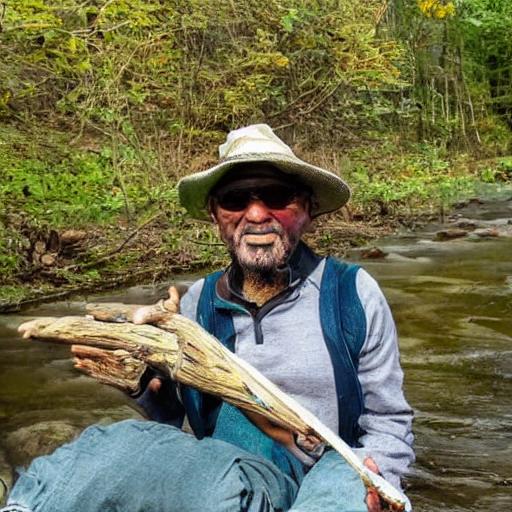}} & a man holding a fish on a river bank. & \raisebox{-0.9\totalheight}{\includegraphics[width=0.1\textwidth]{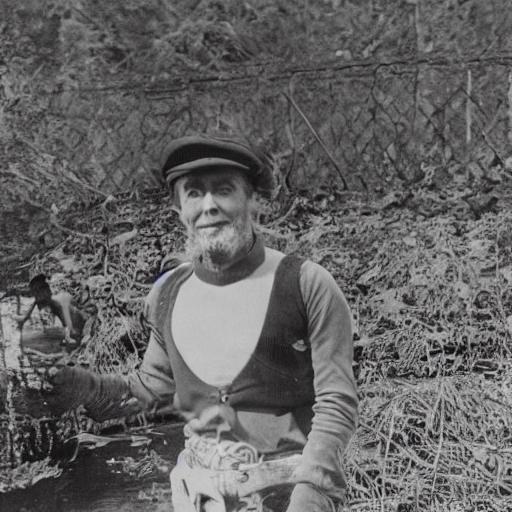}} & a dead fish laying on the ground in the grass & \raisebox{-0.9\totalheight}{\includegraphics[width=0.1\textwidth]{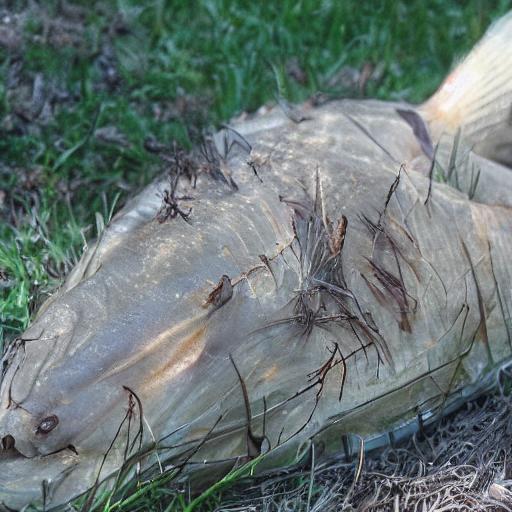}} \\
    \specialrule{2pt}{1\jot}{0.5pc}

    \multirow{12}{*}{\rotatebox{90}{English Springer Spaniel}} &  a brown and white dog with a red collar &  \raisebox{-0.9\totalheight}{\includegraphics[width=0.1\textwidth]{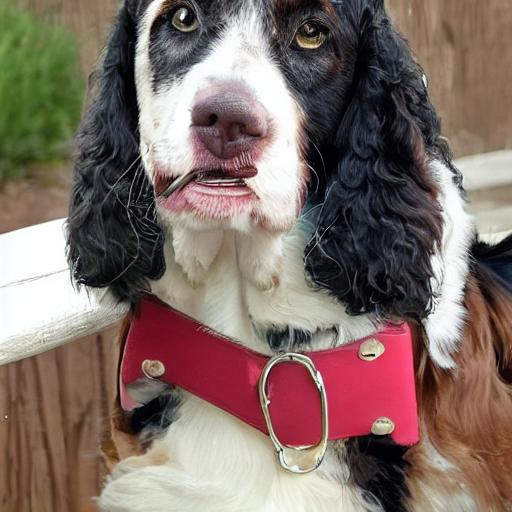}} & a dog sitting on the grass with a leash & \raisebox{-0.9\totalheight}{\includegraphics[width=0.1\textwidth]{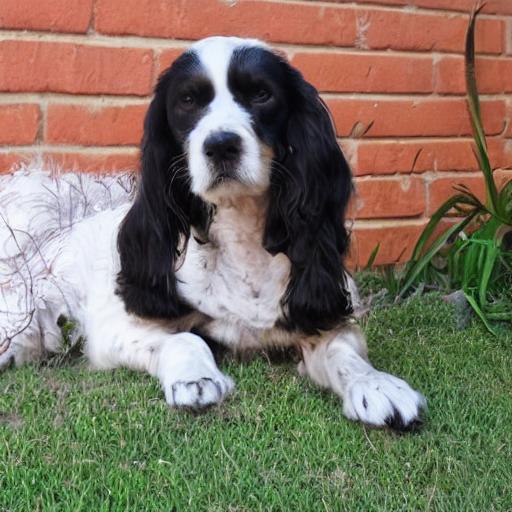}} & a brown and white dog sitting on the ground & \raisebox{-0.9\totalheight}{\includegraphics[width=0.1\textwidth]{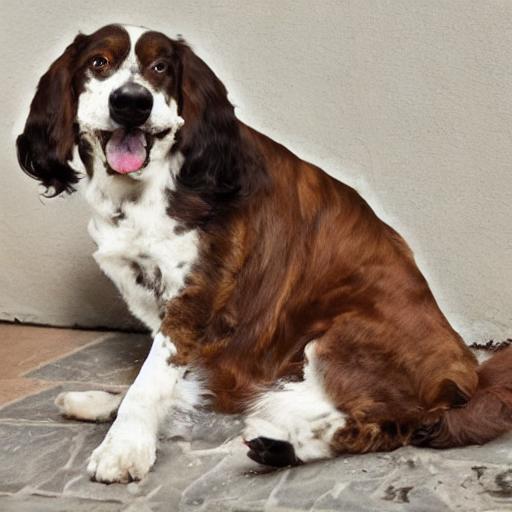}}\\
        & a brown and white dog sitting on a wooden bench &  \raisebox{-0.9\totalheight}{\includegraphics[width=0.1\textwidth]{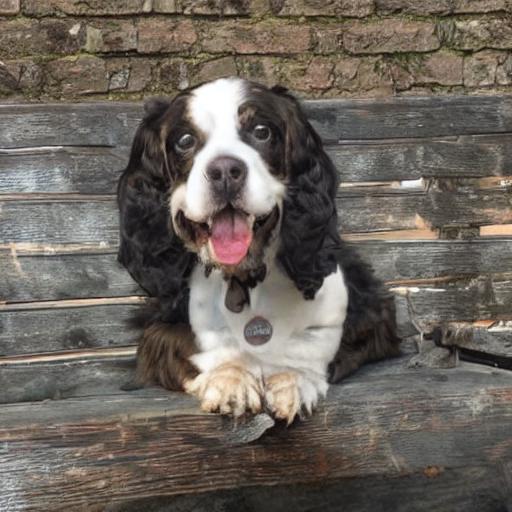}} & a brown and white dog sitting on a wooden bench & \raisebox{-0.9\totalheight}{\includegraphics[width=0.1\textwidth]{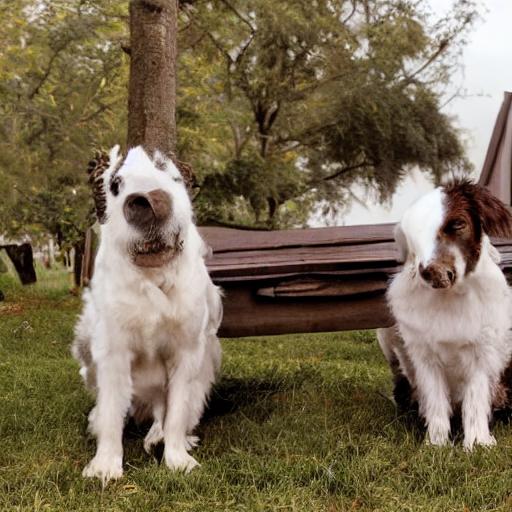}} & a brown and white dog laying on a couch & \raisebox{-0.9\totalheight}{\includegraphics[width=0.1\textwidth]{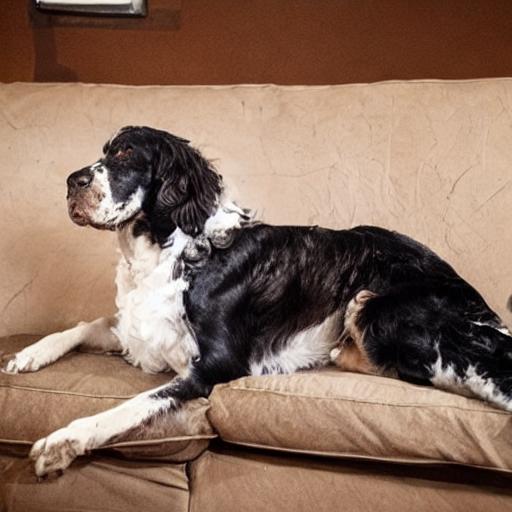}}\\
    & a black and white dog wearing a christmas hat &  \raisebox{-0.9\totalheight}{\includegraphics[width=0.1\textwidth]{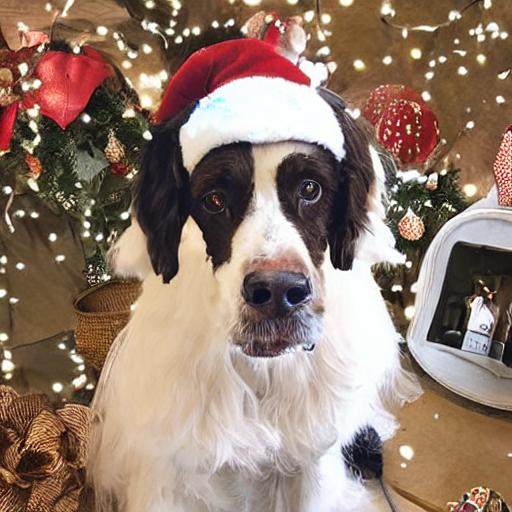}} & a dog wearing a santa hat & \raisebox{-0.9\totalheight}{\includegraphics[width=0.1\textwidth]{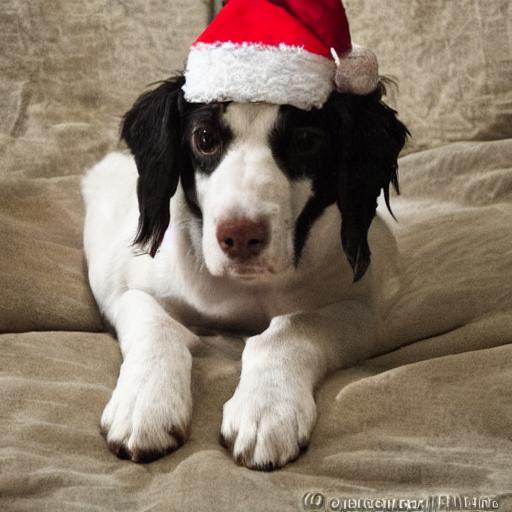}} & aa dog laying on a bed with its head on the pillow & \raisebox{-0.9\totalheight}{\includegraphics[width=0.1\textwidth]{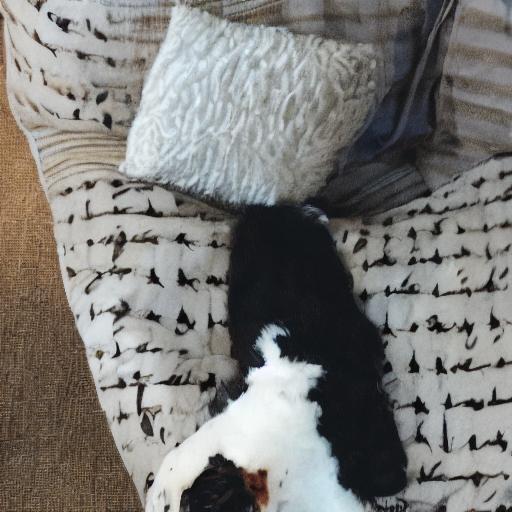}} \\
    \specialrule{2pt}{1\jot}{0.5pc}

    \multirow{12}{*}{\rotatebox{90}{Golf ball}} &  a green and white ball is in the grass &  \raisebox{-0.9\totalheight}{\includegraphics[width=0.1\textwidth]{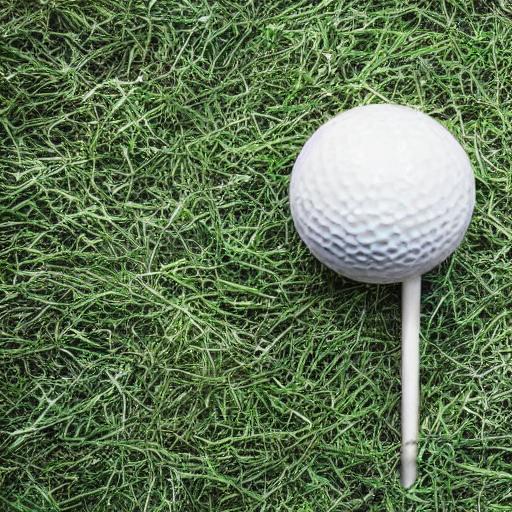}} & a golf ball on a tee with a green background & \raisebox{-0.9\totalheight}{\includegraphics[width=0.1\textwidth]{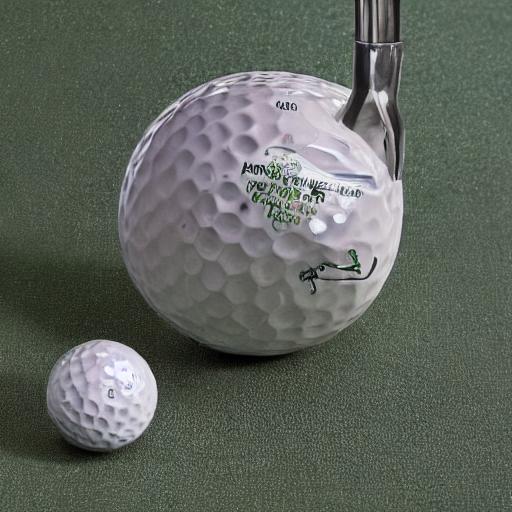}} & a golf ball sits on the green in front of a tree & \raisebox{-0.9\totalheight}{\includegraphics[width=0.1\textwidth]{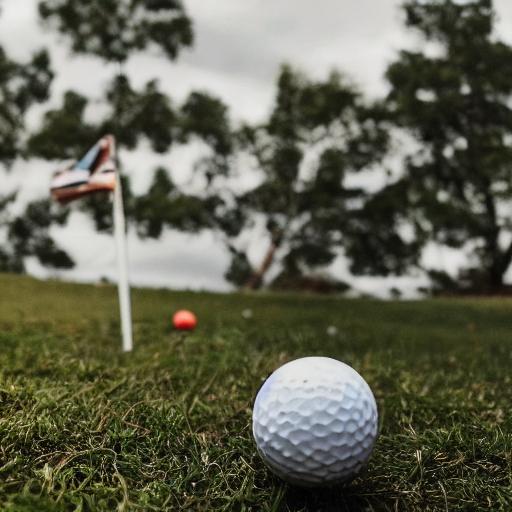}}\\
        & a green and yellow ball is in the grass &  \raisebox{-0.9\totalheight}{\includegraphics[width=0.1\textwidth]{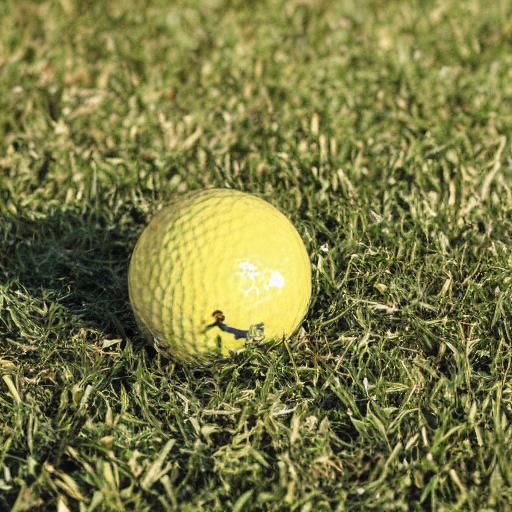}} & golf ball and driver on green grass & \raisebox{-0.9\totalheight}{\includegraphics[width=0.1\textwidth]{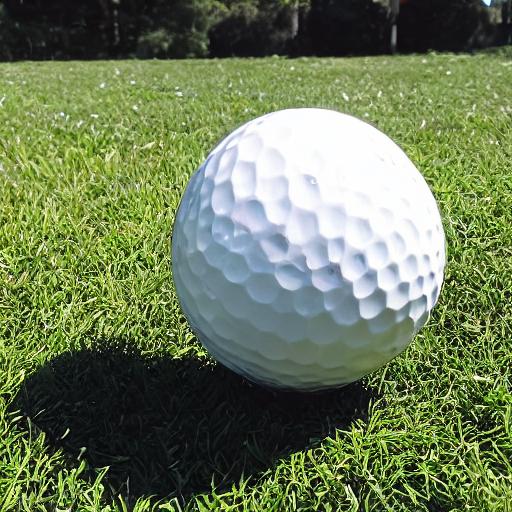}} & a white ball with holes in it on a wall & \raisebox{-0.9\totalheight}{\includegraphics[width=0.1\textwidth]{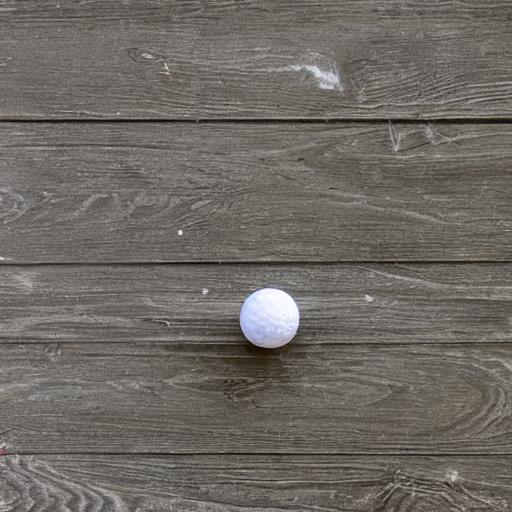}}\\
    & a box filled with different types of balls &  \raisebox{-0.9\totalheight}{\includegraphics[width=0.1\textwidth]{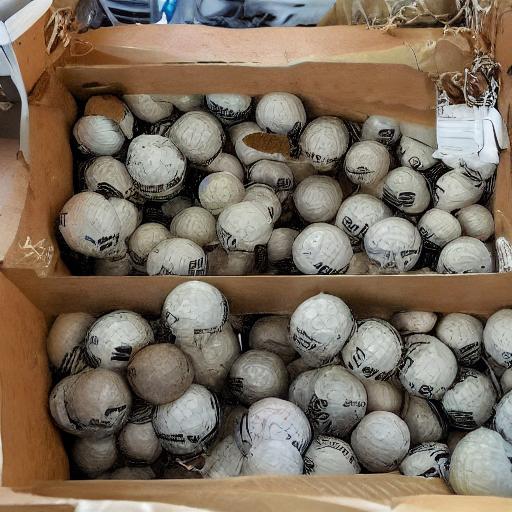}} & a box of golf balls with different logos on them & \raisebox{-0.9\totalheight}{\includegraphics[width=0.1\textwidth]{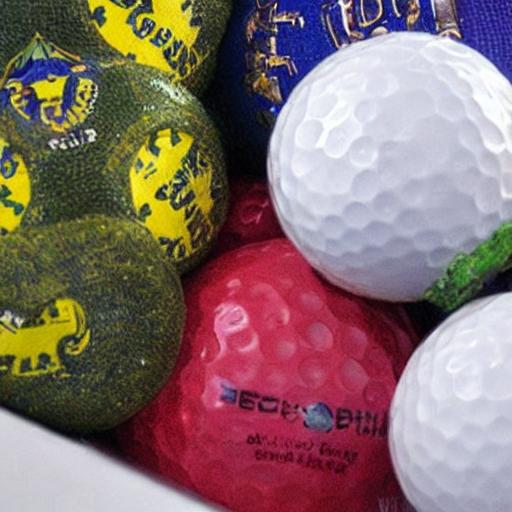}} & a golf ball sits on the grass in a field & \raisebox{-0.9\totalheight}{\includegraphics[width=0.1\textwidth]{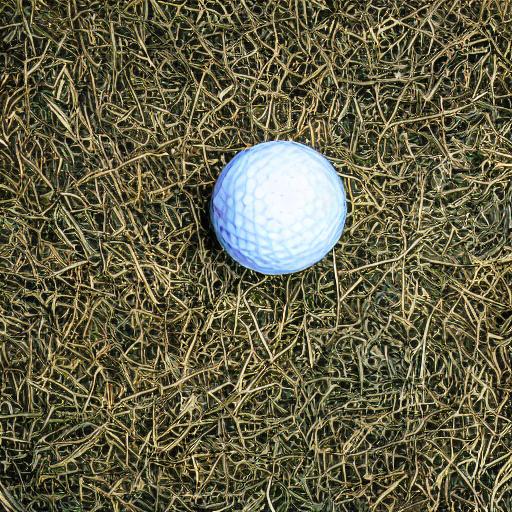}} \\
    \bottomrule
    \end{tabular}
    }
    \label{tab:caption and images}
\end{table*}

\subsection{Examples of Captions}
\label{app:llm rewrited caption}
We present more captions generated by BLIP-2 in Table \ref{tab:caption}. In addition, we show some examples of { (1) Vicuna-generated captions in Table \ref{tab:llm generated caption}, (2) Vicuna-summarized captions in Table \ref{tab:llm summarized caption},} and (3) Vicuna-rewritten captions based on original BLIP-2 captions obtained from real data and zero-shot BLIP-2 captions obtained from preliminary synthetic data in Table \ref{tab:llm caption blip2} and Table \ref{tab:llm caption zero-shot}, respectively.
\begin{table*}[h!]
    \centering
    \caption{Examples of image captions generated by BLIP-2.}
    \begin{tabular}{c|c}
    \toprule
        Class name & Captions \\
    \midrule
        \multirow{5}{*}{Tench} & a fish laying on the grass in the grass \\
         & a man kneeling down holding a large fish \\
         & a large yellow fish swimming in an aquarium \\
         & a fish with a fly rod and a hook \\
         & a young boy in a hoodie holding a fish \\
    \midrule
    \multirow{5}{*}{English Springer} & a dog sitting on the grass with a leash \\
         & a woman is holding a dog on the grass \\
         & two black and white dogs sitting on a couch \\
         & a dog standing in front of a door in a hallway \\
         & a dog holding a toy in its mouth \\
    \midrule
    \multirow{5}{*}{Pirate} & a large boat in the water \\
         & a black and white photo of a boat docked at a dock \\
         & a sailboat in the ocean with a mountain in the background \\
         & a wooden pirate ship \\
         & a large ship docked at a dock with people standing on the dock \\
    \bottomrule
    \multirow{5}{*}{Gibbon} & a monkey standing on a table with a plate of food \\
         & a white monkey hanging from a rope in an enclosure \\
         & a baby gibbon hanging from a tree \\
         & a monkey standing on a table with a plate of food \\
         & a black and white photo of a monkey sitting on a wooden post \\
    \bottomrule
    \multirow{5}{*}{Chainsaw} & a chainsaw with the words ``4020'' on it \\
         & a man in a yellow jacket is cutting down a tree \\
         & a man standing on top of a log with a chainsaw \\
         & two men are cutting down a tree in the fores \\
         & young boy holding a chainsaw in a kitchen \\
    \bottomrule
    \end{tabular}
    \label{tab:caption}
\end{table*}

\newpage

\begin{table*}[h!]
    \centering
    \caption{Examples of image captions generated by Vicuna-13B.}
    \begin{tabular}{c|p{10cm}}
    \toprule
        Class name & Captions \\
    \midrule
        \multirow{8}{*}{Tench} & A photo of tench in a pond, surrounded by lush greenery \\
         & A photo of tench swimming in a river, with a clear blue sky in the background \\
         & A photo of tench in an aquarium, with colorful coral and fish swimming around it \\
         & A photo of tench in a rice field, with a farmer in the background \\
         & A photo of tench in a lake, with a beautiful sunset in the background \\
    \midrule
    \multirow{10}{*}{English Springer}       & A photo of English Springer Spaniel - This breed is known for its friendly and affectionate nature \\
         & A photo of English Springer Spaniel - They are highly intelligent and easy to train \\
         & A photo of English Springer Spaniel - They are great family pets and love to play with children \\
         & A photo of English Springer Spaniel - They have a beautiful and distinctive coat that requires regular grooming \\
         & A photo of English Springer Spaniel - They are versatile hunters and can be used for flushing out game \\
    \midrule
    \multirow{10}{*}{Pirate} & A photo of pirate ship, approaching a rocky coastline with a lighthouse in the distance \\
         & A photo of pirate ship, battling a massive storm with waves crashing over the deck \\
         & A photo of pirate ship, docked at a bustling port with other ships and buildings in the backgroundd \\
         & A photo of pirate ship, sailing through a misty fog with the sun setting in the background \\
         & A photo of pirate ship, sailing towards a group of other pirate ships in a coordinated attack \\
    \bottomrule
    \multirow{5}{*}{Gibbon} & A photo of gibbon sitting on a tree branch with a playful expression \\
         & A photo of gibbon swinging from tree to tree in the jungle \\
         & A photo of gibbon making a loud call in the jungle \\
         & A photo of gibbon eating fruit in a tree \\
         & A photo of gibbon playing with a ball of leaves \\
    \bottomrule
    \multirow{8}{*}{Chainsaw} & A photo of chainsaw - A chainsaw is an essential tool for any lumberjack or tree surgeon \\
         & A photo of chainsaw - This chainsaw is perfect for cutting through large logs and branches \\
         & A photo of chainsaw - This chainsaw is designed for heavy-duty use and can handle even the toughest jobs \\
         & A photo of chainsaw - This chainsaw is equipped with a powerful engine that makes cutting through wood a breeze \\
    \bottomrule
    \end{tabular}
    \label{tab:llm generated caption}
\end{table*}

\newpage

\begin{table*}[h!]
    \centering
    \caption{Examples of image captions summarized by Vicuna-13B from real image captions.}
    \begin{tabular}{c|p{10cm}}
    \toprule
        Class name & Captions \\
    \midrule
        \multirow{5}{*}{Tench} & A photo of tench: A fish laying on the grass in the grass \\
         & A photo of tench: A man kneeling down holding a large fish \\
         & A photo of tench: A man holding a fish on a river bank \\
         & A photo of tench: A man in camouflage holding a fish at night \\
         & A photo of tench: A man holding a large fish in the dark \\
    \midrule
    \multirow{8}{*}{English Springer} & A photo of an English Springer Spaniel laying on a bed with a yellow toy \\
         & A photo of English Springer Spaniel standing in the grass with its tongue out \\
         & A photo of English Springer Spaniel with a toothbrush in its mouth \\
         & A photo of English Springer Spaniel running through a field with a ball in its mouth \\
         & A photo of an English Springer Spaniel with a blue collar \\
    \midrule
    \multirow{5}{*}{Pirate} & A photo of pirate ship: A ship in the water \\
         & A photo of pirate ship: A group of people dressed in pirate costumes \\
         & A photo of pirate ship: A pirate ship made out of lego bricks \\
         & A photo of a pirate ship docked in the ocean \\
         & A photo of a red and black pirate ship \\
    \bottomrule
    \multirow{5}{*}{Gibbon}    & A photo of gibbon sitting on a tree branch with its arms crossed \\
         & A photo of gibbon sitting on a rock in a zoo \\
         & A photo of gibbon in a tree \\
         & A photo of gibbon in a cage \\
         & A photo of gibbon on a wooden fence \\
    \bottomrule
    \multirow{5}{*}{Chainsaw} & A photo of an ECHO chainsaw \\
         & A photo of a chainsaw in front of a world map \\
         & A photo of a green chainsaw sitting on top of a wooden box \\
         & A photo of a chainsaw with a green and red handle \\
         & A photo of a man holding a chainsaw in the desert \\
    \bottomrule
    \end{tabular}
    \label{tab:llm summarized caption}
\end{table*}

\newpage

\begin{table*}[h!]
    \centering
    \caption{Vicuna-rewritten captions based on BLIP-2 captions on real data.}
    \begin{tabular}{c|p{10cm}}
    \toprule
        Class name & \multicolumn{1}{c}{Captions} \\
    \midrule
        \multirow{14}{*}{Tench} & A man proudly displays a caught tench fish on the grass, surrounded by nature.\\
        \cmidrule{2-2}
         & A heartwarming scene of a man and his beloved dog, celebrating a memorable fishing trip, with the angler triumphantly holding a tench and surrounded by the captivating natural scenery. \\
         \cmidrule{2-2}
         & A fisherman, surrounded by lush vegetation, poses with a tench he caught in the river. \\
         \cmidrule{2-2}
         & Tench are a hardy fish species and can survive in a wide range of water temperatures and conditions. \\
         \cmidrule{2-2}
         & Tench are a popular food fish in many parts of the world. This image shows a fisherman holding three tench caught in a river, ready to be cleaned and cooked. \\
    \midrule
    \multirow{10}{*}{English Springer Spaniel} & A group of English Springer Spaniels running in a park \\
    \cmidrule{2-2}
         & This English Springer Spaniel is casually sitting in front of a row of crates, seemingly keeping an eye on them. \\
         \cmidrule{2-2}
         & A cute English Springer Spaniel eagerly waiting for its owner to throw a toy, while sitting on a wooden deck. \\
         \cmidrule{2-2}
         & English Springer Spaniel puppy exploring the world, sniffing the flowers and chasing butterflies in a sunny meadow. \\
         \cmidrule{2-2}
         & An English Springer Spaniel happily swimming in a lake, with a small boat moored nearby. \\
    \midrule
    \multirow{13}{*}{Golf ball} & A mouthwatering golf ball cake, displayed on a sleek glass pedestal with a silver base, adding a touch of elegance to any event. \\
    \cmidrule{2-2}
         & A professional golfer prepares for a major tournament, practicing their swing and focusing on their mental game to stay focused and calm under pressure. \\
         \cmidrule{2-2}
         & Golf ball display: A golf shop assistant is arranging a display of different golf balls, showcasing their vibrant colors and unique features to attract potential customers. \\
         \cmidrule{2-2}
         & Double Trouble: Two golf balls are shown, each with its own trail of smoke, suggesting a doubles match with each player aiming for the hole. \\
         \cmidrule{2-2}
         & A cupcake with a golf ball on top, placed on a chocolate brown background with white polka dots. \\
    \bottomrule
    \end{tabular}
    \label{tab:llm caption blip2}
\end{table*}

\begin{table*}[ht]
    \centering
    \caption{Vicuna rewritten captions based on zero-shot BLIP-2 captions on preliminary synthetic data.}
    \begin{tabular}{c|p{10cm}}
    \toprule
        Class name & \multicolumn{1}{c}{Captions} \\
    \midrule
        \multirow{13}{*}{Tench} & A cook preparing to cook a tench fish on a grill, with various vegetables and herbs ready to be added. \\
        \cmidrule{2-2}
         & A lifeless tench rests on a bed of aquatic vegetation, its body perpendicular to the water's surface, as an ominous overcast sky looms above.\\
         \cmidrule{2-2}
         & A group of friends enjoying a picnic by a lake, with a tench fish on a plate in front of them. They are smiling and laughing, and the fish is the centerpiece of their meal. \\
         \cmidrule{2-2}
         & A nostalgic image of a man wearing a classic suit and tie, paying homage to the timeless elegance of the past. \\
         \cmidrule{2-2}
         & Tench is a versatile ingredient in various cuisines, such as French, Italian, and Japanese. \\
    \midrule
    \multirow{15}{*}{English Springer Spaniel} & An English Springer Spaniel with a wagging tail is sitting on a grassy lawn, in front of a charming cottage, and looking up at its owner. \\
    \cmidrule{2-2}
         & This fluffy English Springer Spaniel is the perfect friend for any adventure! With its energetic and friendly demeanor, it's always ready to explore new trails or catch a Frisbee in the park. \\
         \cmidrule{2-2}
         & A well-behaved English Springer Spaniel, alert and curious, sitting in a cozy basket made of red and white stripes. It has a black and white coat that complements the color scheme of the basket. The dog's intelligent eyes look around, eager to explore and learn. \\
         \cmidrule{2-2}
         & Nature's beauty, as seen through the eyes of a loyal companion. \\
         \cmidrule{2-2}
         & A portrait of an English Springer Spaniel by Julie Mcclure, where the dog is shown sitting on a wooden dock, overlooking a calm and serene lake, with a peaceful and contemplative expression, showcasing the dog's intelligence and independent nature. \\
    \midrule
    \multirow{12}{*}{Golf ball} & A white golf ball rolling down a lush green hill, chased by a determined golfer. \\
    \cmidrule{2-2}
         & A shiny golf ball perches on the tee, as a sleek golf club lies partially hidden in the verdant grass. \\
         \cmidrule{2-2}
         & A moment of stillness captures a golf ball perched atop a gravel bed, surrounded by the hustle and bustle of the course. \\
         \cmidrule{2-2}
         & A golden retriever playfully jumps and pounces on a golf ball, sending it bouncing across the grass in a game of fetch. \\
         \cmidrule{2-2}
         & A bird with vibrant plumage hovers above a bird feeder, its keen eyes scanning the surrounding meadow for any signs of prey. \\
    \bottomrule
    \end{tabular}
    \label{tab:llm caption zero-shot}
\end{table*}